\documentclass[letterpaper, 10 pt, conference]{ieeeconf}  

\IEEEoverridecommandlockouts                              

\usepackage{soul}
\usepackage{xcolor}
\usepackage{amsmath}
\usepackage[cal=pxtx]{mathalfa} 
\usepackage{graphicx}
\usepackage{amsfonts}
\usepackage{subcaption}
\usepackage{multirow}
\usepackage{float}
\usepackage{stfloats}
\usepackage{url}
\usepackage{algorithm}
\usepackage{algorithmicx}  
\usepackage{algpseudocode}
\usepackage{lscape}
\usepackage{pdflscape}
\usepackage{adjustbox}   
\usepackage{fancyhdr}

\newcommand{\icraMatthew}[1]{{\color{black} #1}}
\newcommand{\icra}[1]{{\color{black} #1}}
\newcommand{\RAL}[1]{{\color{black} #1}}
\newcommand{\RALedit}[1]{{\color{black} #1}}

\title{\LARGE \bf
Faster Model Predictive Control via Self-Supervised Initialization Learning
}

\author{Zhaoxin Li\textsuperscript{* 1}, 
Xiaoke Wang\textsuperscript{* 1}, 
Letian Chen\textsuperscript{1}, 
Rohan Paleja\textsuperscript{2}, 
Subramanya Nageshrao\textsuperscript{3},
Matthew Gombolay\textsuperscript{1} 
\thanks{1. Georgia Institute of Technology, Atlanta, GA 30332, USA}%
\thanks{2. MIT Lincoln laboratory, Lexington, MA 02421}%
\thanks{3. Ford Motor Company, Dearborn, MI 48120, USA}%
}

\begin{document}

\fancypagestyle{firstpage}{
  \fancyhf{}
  \fancyfoot[C]{\footnotesize This work has been submitted to the IEEE for possible publication. Copyright may be transferred without notice, after which this version may no longer be accessible.}
  \renewcommand{\headrulewidth}{0pt}
  \renewcommand{\footrulewidth}{0pt}
}

\maketitle
\thispagestyle{firstpage}
\pagestyle{empty}

\begin{abstract}

Model Predictive Control (MPC) is widely used in robot control by optimizing a sequence of control outputs over a finite-horizon. Computational approaches for MPC include deterministic methods (e.g., iLQR and COBYLA), as well as sampling-based methods (e.g., MPPI and CEM). However, complex system dynamics and non-convex or non-differentiable cost terms often lead to prohibitive optimization times that limit real-world deployment. Prior efforts to accelerate MPC have limitations on: (i) reusing previous solutions fails under sharp state changes and (ii) pure imitation learning does not target compute efficiency directly and suffers from suboptimality in the training data. To address these, We propose a warm-start framework that learns a policy to generate high-quality initial guesses for MPC solver. The policy is first trained via behavior cloning from expert MPC rollouts and then fine-tuned online with reinforcement learning to directly minimize MPC optimization time. We empirically validate that our approach improves both deterministic and sampling-based MPC methods, achieving up to 21.6\% faster optimization and 34.1\% more tracking accuracy for deterministic MPC in Formula 1 track path-tracking domain, and improving safety by 100\%, path efficiency by 12.8\%, and steering smoothness by 7.2\% for sampling-based MPC in obstacle-rich navigation domain. These results demonstrate that our framework not only accelerates MPC but also improves overall control performance. Furthermore, it can be applied to a broader range of control algorithms that benefit from good initial guesses.


\end{abstract}

\section{INTRODUCTION}
\icraMatthew{Algorithms that optimize control outputs iteratively like MPC have been widely adopted to control dynamic systems}, such as autonomous vehicles~\cite{borrelli2005mpc, karnchanachari2020practical, kong2015kinematic, kim2022smooth}, aircraft~\cite{bauersfeld2021mpc, jadbabaie2002control}, humanoid robots~\cite{kuindersma2016optimization}, etc. \RAL{While gradient-based MPC is highly effective when the dynamics and cost functions are smooth and differentiable, applying MPC in real-world settings remains challenging. Computational bottlenecks often arise in systems with non-convex or non-differentiable costs and long planning horizons, where gradient-free solvers are typically employed~\cite{bouzidi2023learning, lembono2020learning, mansard2018using, richter2009real}.}

\begin{figure}
  \centering
  \includegraphics[width= 0.46\textwidth]{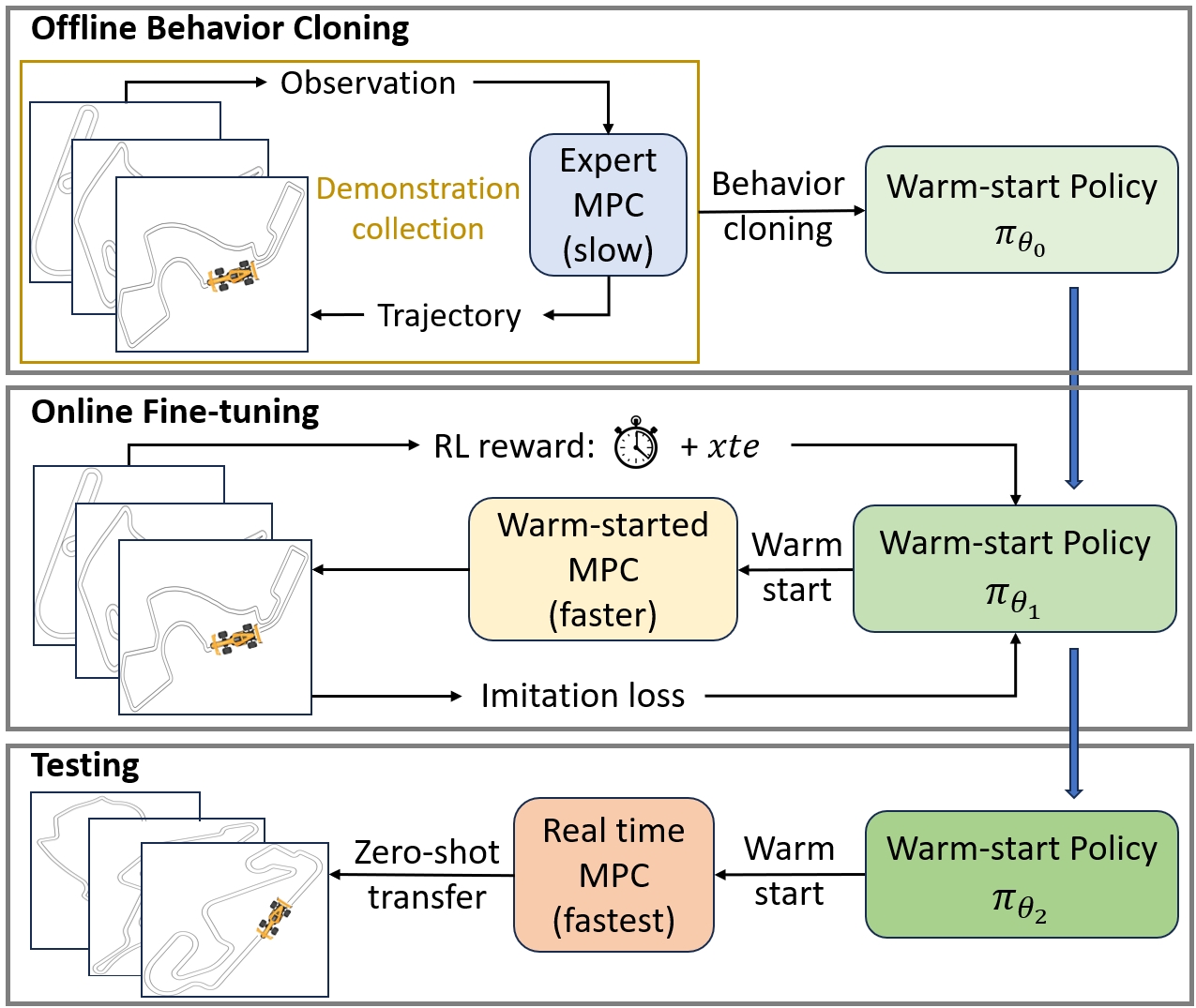}
  \caption{Overview of our proposed algorithm. The first two blocks denote the two-phase training framework. In the first phase, we collect expert MPC demonstrations and train a warm-start policy using behavior cloning to speed up MPC. In the second phase, we fine-tune this policy within an online training framework to enhance its performance and generalizability. During testing, the proposed framework is evaluated on both training tracks and challenging zero-shot tracks, as demonstrated in the third block.}
  \label{fig:general_framework}
  \vspace{-2em}
\end{figure}

\RALedit{Dynamic replanning in MPC requires fast computation to operate in real time, but standard MPC without warm-starting often fails to achieve sufficient speed in challenging domains~\cite{zeilinger2011real, balliu2025exponential}. Warm starting addresses this by initializing the solver with a solution close to the optimum, reducing the search space and accelerating convergence.} 
In this paper, we choose to test our proposed algorithm to warm start Model Predictive Control (MPC), which optimizes a trajectory over a predefined horizon.

\RALedit{Prior works have employed heuristic strategies to warm-start MPC, but these methods often struggle with scalability and fail to adapt effectively to sudden changes in system states.}
Traditionally, one common technique involves utilizing the MPC solution from the previous sampling instance as the initial guess for the current control step~\cite{zeilinger2011real, pan2020imitation}. However, this approach falls short when faced with sudden state changes (e.g., the vehicle approaches a sharp turn). Another method involves maintaining a memory buffer to store \RALedit{and leverage} historical MPC solution~\cite{mansard2018using, marcucci2020warm}. \RALedit{However,} this approach fails to generalize to unseen states and scales.

\RALedit{Traditional machine learning approach has also been adopted to learn a warm-start for MPC, however, it suffers from inaccurate initialization. }
Klaučo et al. propose using a k-NN classifier to classify the solution space into active sets for the solver to search for a solution~\cite{klauvco2019machine}, \RAL{but this only produces coarse guesses. These shortcomings highlight the need for a general warm-starting framework that directly learns to generate high-quality initial guesses, scales to complex domains, and improves solver efficiency under sharp dynamics and \RALedit{diverse} conditions.}


\RALedit{More recent approaches have attempted to predict effective warm-starts for model predictive optimization by either behavior cloning (BC) historical trajectories~\cite{kusumoto2019informed, sacks2023learning} or training reinforcement learning (RL) policies for trajectory generation~\cite{sacks2024deep}. While BC is fast and straightforward to train, it cannot directly learn to minimize the optimization time of MPC. Instead, all supervised learning can do is to imitate the output of the solver. Worse yet, BC suffers from covariance shift, leading to poor generalization in zero-shot scenarios~\cite{nado2020evaluating}. In contrast, RL is capable of directly optimizing performance objectives but requires extensive interaction data and struggles to learn from scratch~\cite{li2017deep}. As such, there remains a need to bridge these two paradigms—combining the efficiency of BC with the adaptability of RL—to effectively warm-start MPC across diverse solver families.}

\RALedit{To overcome the limitations of prior work, we propose a learning framework that trains a policy to generate high-quality initial guesses for optimization-based control algorithms (e.g., MPC).}
\RAL{The framework combines offline behavior cloning (BC) with online reinforcement learning (RL) fine-tuning and an additional zero-shot testing phase, as depicted in Fig.~\ref{fig:general_framework}. The two training phases work synergistically. The online-fine tuning phase uses RL to address the suboptimality and covariance shift problem of the offline BC by directly optimizing the optimization time of MPC. \RALedit{At the same time, the offline BC} provides the online learning algorithm with a good starting point to expedite the training process.} Moreover, initializing MPC with a learned policy rather than replacing MPC with an end-to-end controller preserves the original control formulation. The policy only sets the starting point, while MPC enforces system constraints to ensure safer operation. \RALedit{We demonstrate the generality of our approach by evaluating it across both deterministic MPC solver (COBYLA~\cite{powell1994direct}) and sampling-based MPC solver (MPPI~\cite{williams2015model}). Our key contributions are: }

\begin{enumerate}
    \item \RALedit{We propose a novel two-phase learning framework to warm-start optimization-based control algorithms (e.g., MPC), enhancing both efficiency and robustness.} 
    \item \RAL{We empirically evaluate the framework on high-speed vehicle control tasks (\RALedit{deterministic MPC}), achieving $21.6\%$ faster optimization and $34.1\%$ higher tracking accuracy on zero-shot Formula 1 tracks compared to single-phase training.} 
    \item \RAL{We further test the framework on path-planning tasks (sampling-based MPC), showing $100\%$ improved safety, $12.8\%$ higher path efficiency, and $7.2\%$ smoother steering on zero-shot tracks compared to single-phase training.} 
\end{enumerate}

\section{\RALedit{Preliminaries - Model Predictive Control}}



\RALedit{MPC is formulated as a finite-horizon optimization problem that minimizes a cost function while satisfying system dynamics and constraints. The MPC control law is formulated in Eqs.\eqref{eq:mpc:nominal}. At each time step $t$, MPC observes the current state $x_t$ and computes an optimal sequence of control inputs $U = [u_t, \ldots, u_{t+H-1}]$ over a planning horizon $H$ by minimizing a cost function $J$ (Eq.~\eqref{eq:mpc:nominal1}). The system dynamics $x_{t+i+1} = f(x_{t+i}, u_{t+i})$ (Eq.~\eqref{eq:mpc:nominal2}) predict how control inputs influence future states. Eq.~\eqref{eq:mpc:nominal3} and Eq.~\eqref{eq:mpc:nominal4} specify the constraints for the control inputs and states. After solving the optimization, only the first control input $u_t$ is applied, and the process repeats at the next time step using the updated state.}
\vspace{-1.5em}

\begin{subequations}
\label{eq:mpc:nominal}
\begin{gather}
    J = \underset{U}{\text{minimize}} \sum_{i=0}^{H-1} l(x_{t+i}, u_{t+i}) \label{eq:mpc:nominal1} \\
    \text{subject to} \quad x_{t+i+1} = f(x_{t+i}, u_{t+i}) \label{eq:mpc:nominal2} \\
    U=[u_t, ..., u_{t+H-1}] \in \mathcal{U}_j \ \text{for all} \ j=1,\dots,n_{c_u} \label{eq:mpc:nominal3} \\
    X=[x_t, ..., x_{t+H}] \in \mathcal{X}_j \ \text{for all} \ j=1,\dots,n_{c_x} \label{eq:mpc:nominal4}
\end{gather}
\end{subequations}


\section{Method}

In this section, we discuss in detail the proposed two-phase training framework \RAL{to learn a warm-start policy that reduces MPC solver runtime while maintaining tracking performance.}. In the first phase, we run the MPC to collect the expert demonstrations, which are represented as state-action pairs. Then, we use BC to train a warm-start policy to mimic the expert MPC's solution, as shown in Algorithm~\ref{alg:offline}. The output of warm-start policy is utilized as an initial guess to warm start the MPC. In the second phase, we load the pre-trained trajectory prediction model into an online training framework and fine tune the warm-start policy to address the suboptimality problem caused by BC and improve the model's generalizability. The online fine tuning phase is shown in Algorithm~\ref{alg:online}.

\begin{algorithm}
  \caption{Offline Training}
  \label{alg:offline}
  \begin{algorithmic}[1]
    \State \textbf{Input:} expert MPC $\pi^\text{MPC}$ with planning horizon $H$ and maximum optimization iteration $N_{expert}$ and all-zero vector $\vec{0}$ as initial guess, environment transition $T$, number of state-action pairs to collect $N$
    \State Initialize neural network policy $\pi_\theta^\text{warm}$
    \State $t\leftarrow 0, s\leftarrow s_0, \mathcal{D}=\emptyset$
    \While{$t < N$}
    \State $(u_t,u_{t+1},\cdots,u_{t+H-1})\leftarrow \pi^\text{MPC}(s, \vec{0})$
    \State $\mathcal{D}\leftarrow \mathcal{D}\cup \{(s,u_t,u_{t+1},\cdots,u_{t+H-1})\}$
    \State $s\leftarrow T(s,u_t)$
    \State $t\leftarrow t+1$
    \EndWhile
    \State Train $\pi_\theta^{warm}$ with Eq.~\eqref{eq:BC} and $\mathcal{D}$
  \end{algorithmic}
\end{algorithm}

\vspace{-1em}
\subsection{Offline Training}
\label{sec:offline_training}

At this phase, we implement an expert MPC to collect a dataset $\mathcal{D}$ containing $N$ state-action pairs. The expert MPC, $\pi^{MPC}$, controls the agent to complete the task without a warm-started initial guess. At each step, $\pi^{MPC}$ observes a state $s$, takes an all-zero vector as the initial guess, and optimizes it to output $(u_t,u_{t+1},\cdots,u_{t+H-1})$ (line 5). The state-action pair $(s,u_t,u_{t+1},\cdots,u_{t+H-1})$ is stored in $\mathcal{D}$ (line 6). The first action is then applied, transitioning the system to the next state based on environment transition $T$ (line 7–8). During data collection, $\pi^{MPC}$ runs enough iterations to ensure high control performance by disabling the “early stop.”

We design our warm-start policy, $\pi_{\theta}^{warm}$, as a multi-layer perceptron with ReLU activation function~\cite{glorot2010understanding}. Given the current vehicle state, the $\pi_{\theta}^{warm}$ predicts a sequence of actions that serves as the initial guess of the MPC to warm start the optimization process. In the offline training phase, we utilize BC to train $\pi_{\theta}^{warm}$ (line 10), where BC learns a control policy $\pi_\theta$ from pre-collected MPC demonstrations by minimizing the mean squared error (MSE) between predicted and demonstrated actions, as shown in Eq.~\eqref{eq:BC}.
\vspace{-1em}

\begin{align}
    \theta^*=\min_{\theta}{\sum_{\tau^i\in\mathcal{D}}\sum_{t=0}^T{(\pi_\theta^{warm}(s^i)_t-u_t^i)^2}}
    \label{eq:BC}
\end{align}


\begin{algorithm}
  \caption{Online Fine Tuning}
  \label{alg:online}
  \begin{algorithmic}[1]
    \State \textbf{Input:} fast MPC $\pi^\text{MPC}_\text{fast}$ with planning horizon $H$ and maximum optimization iteration $N_{fast}$, environment transition $T$, pre-trained warm-start policy $\pi_\theta^{warm}$
    \For{each RL training iteration}
      \State Perceive an observation $s$
      \State $\hat{U}_t = (\hat{u}_t, \hat{u}_{t+1}, ..., \hat{u}_{t+H-1}) \leftarrow \pi^{warm}(s)$
      \State $pos_{i+1}^{car} = M_{dynamics}((s_i, \hat{u}_i)) | _{i=t}^{H-1}$ 
      \State Evaluate the quality of the planned trajectory
      \State $U_t^\text{MPC} = (u_t, u_{t+1}, ..., u_{t+H-1}) \leftarrow \pi^\text{MPC}_\text{fast}(s, \hat{U}_t)$

      
      \State Calculate reward in RL using Eq.~\eqref{eq:rl:reward}
      \State $L_{imitation} = MSE(U_t^{MPC}, U_t^{guess})$ 
      \State Compute training loss $L$ in Eq.~\eqref{eq:online_loss} 
      \State $s \leftarrow T(s,u_t)$
      \State Update $\pi_\theta^{warm}$ with $L$
    \EndFor
  \end{algorithmic}
\end{algorithm}

\subsection{Online Fine Tuning}
\label{sec:online_fine_tuning}

\RAL{Our second phase, online fine-tuning, addresses the shortcomings of BC by combining the strengths of RL and Dataset Aggregation (DAgger)~\cite{ross2011reduction}. RL enables us to directly optimize the MPC computation time. Meanwhile, DAgger mitigates the distribution shift problem by augmenting the training set with additional expert demonstrations.}

DAgger builds upon BC by incorporating online interaction with the environment and online querying of the expert. Unlike BC, which trains solely on a fixed dataset of expert demonstrations, DAgger actively collects data from interactions with the environment and solicits expert feedback to augment its training. This online improvement process allows DAgger to learn from a more diverse set of experiences, adapt to new situations, and refine the agent's policy over time, ultimately leading to improved performance in imitation learning tasks.

RL operates under the formalization of Markov Decision Process (MDP), $\mathcal{M}=\langle\mathcal{S}, \mathcal{A}, R, T, \gamma, \rho_0\rangle$. $\mathcal{S}$ is the state space and $\mathcal{A}$ denotes the action space. $R$ encodes the reward of a given state. $T$ is a deterministic transition function that decides the next state, $s'$, when applying the action, $a \in A$, in state, $s \in S$. $\gamma \in (0, 1)$ is the temporal discount factor. $\rho_0$ denotes the initial state probability distribution. A policy, $\pi: S \rightarrow A$
, is a mapping from states to actions or to a probability distribution over actions. The objective of RL is to find the policy that optimizes the expected discounted return, $\pi^* = \mathbb{E}_{\tau \sim \pi}\left[\sum^\infty_{t=0}\gamma^t R(s_t)\right]$. 

DAgger is integrated as a term in actor loss in the RL training, as shown in Eq.~\eqref{eq:online_loss}. $B$ is the batch size and $Q_{\theta}$ is the critic network. $L_{imitation}$ is the loss signal from DAgger, representing MSE between the expert MPC's solution and the warm-started initial guess output from $\pi_{\theta}^{warm}$. DAgger loss is only added when the quality of the initial guess is lower than a certain threshold. \RAL{To reduce the overhead of repeatedly querying the expert MPC, $L_{imitation}$ is only applied when the initial guess falls below a certain threshold. This accelerates RL training while maintaining regularization from expert demonstrations.} 
$\lambda$ is the weight coefficient between the RL loss and imitation loss.

\begin{equation}\label{eq:online_loss}
\begin{aligned}
L & = \lambda \cdot L_{actor} + (1 - \lambda) \cdot L_{imitation} \\
L_{\text{actor}} & = -\frac{1}{B} \sum_{i=1}^{B} Q_{\theta}(s_i, \pi^\text{warm}(s_i))
\cdot L_{value}
\end{aligned}
\end{equation}



\section{EXPERIMENTS SETUP}


In this section, we discuss in detail the experimental setup for both deterministic and sampling-based MPC. For deterministic MPC, we evaluate the path-tracking problem on high-speed Formula 1 tracks~\cite{f1tenth_racetracks}, \RAL{where minimizing optimization time is critical for maintaining performance.} For sampling-based MPC, we choose the navigation problem in environments without a pre-defined reference path, requiring the vehicle to generate its own collision-free trajectory.

\subsection{\RALedit{Deterministic} MPC}

\subsubsection{MPC Details}

\begin{figure}
    \centering
    \begin{subfigure}[b]{0.48\textwidth}
        \centering
        \includegraphics[width=\textwidth]{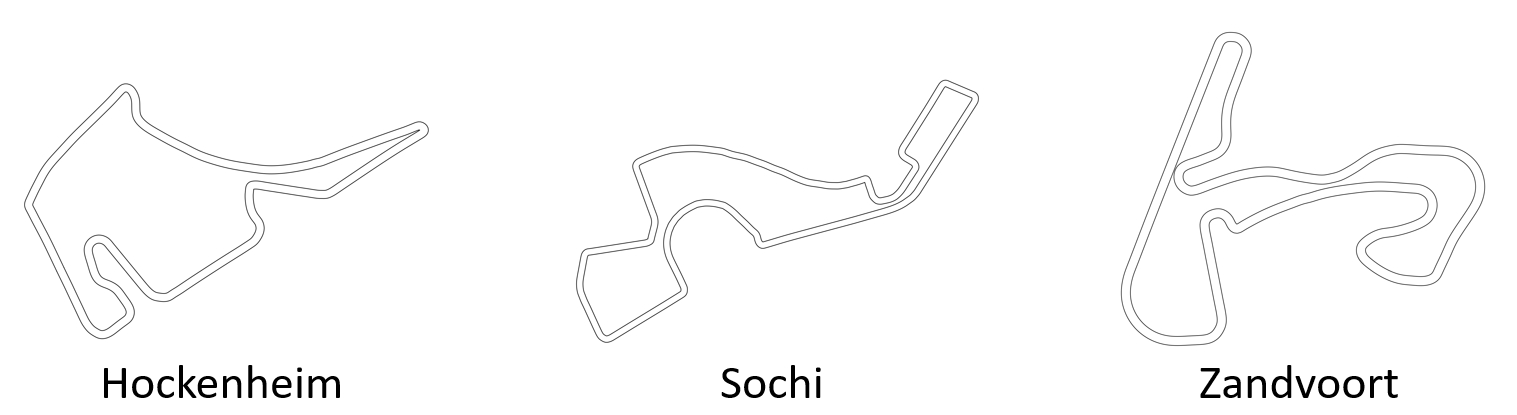}
        \caption{Training tracks}
        \label{fig:training_tracks}
    \end{subfigure}
    \hfill
    \begin{subfigure}[b]{0.48\textwidth}
        \centering
        \includegraphics[width=\textwidth]{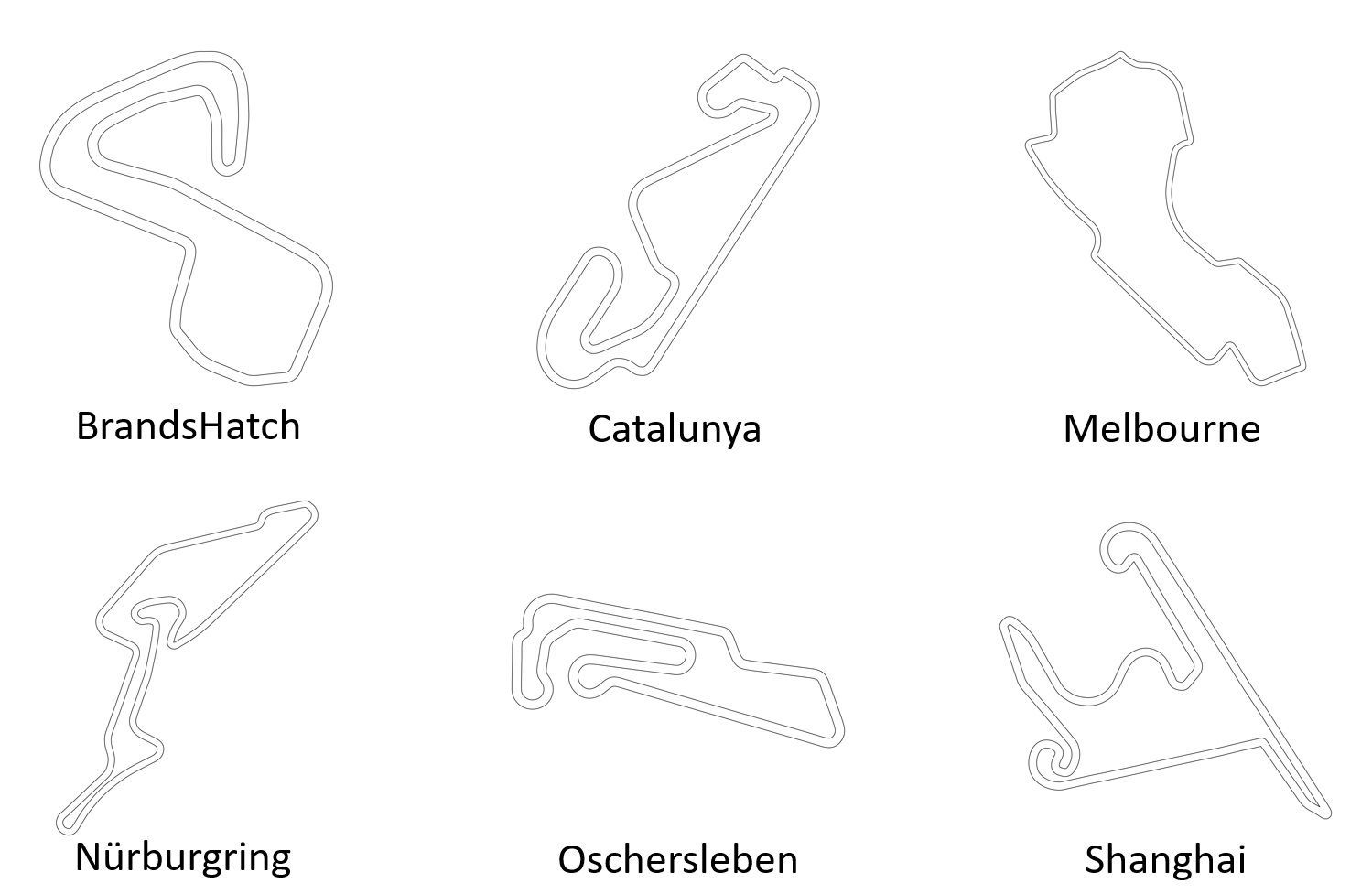}
        \caption{Zero-shot tracks}
        \label{fig:zero_shot_tracks_complex}
    \end{subfigure}
    \caption{\RAL{Training and Testing maps used in the Formula 1 track path-tracking domain.}}
    \label{fig:f1_tracks}
    \vspace{-1em} 
\end{figure}

The Formula 1 domain consists of three training and seven zero-shot testing tracks, as shown in Fig.~\ref{fig:training_tracks} and Fig.~\ref{fig:zero_shot_tracks_complex}. The training tracks are used for demonstration collection and policy learning, while the zero-shot tracks remain unseen during training to test generalization. Each track’s reference trajectory is defined by waypoints along the centerline and scaled down 10:1 to make each lap a reasonable length. The friction between the tire and the road is not considered in our dynamics model.

\vspace{-1em}

\begin{subequations}
\label{eq:mpc:dynamic_model}
\begin{gather}
    x_{t+1}^{car} = x_t^{car} + v_t \cdot \cos(yaw_t) \cdot dt + \epsilon_1\\
    y_{t+1}^{car} = y_t^{car} + v_t \cdot \sin(yaw_t) \cdot dt + \epsilon_2 \\
    yaw_{t+1} = yaw_t + \frac{v_t}{L} \cdot \tan(\theta_t^{\text{steering}}) \cdot dt + \epsilon_3 \\
    v_{t+1} =
    \begin{cases}
        v_t + a_t \,\Delta t + \epsilon_4, & \text{if } v_t + a_t \,\Delta t < 10, \\
        10, & \text{otherwise.}
    \end{cases} \\
    \boldsymbol{\epsilon} \sim \mathcal{N}(\mathbf{0}, \sigma^2 I),
    \quad \boldsymbol{\epsilon} = [\epsilon_1, \epsilon_2, \epsilon_3, \epsilon_4]^\top 
\end{gather}
\end{subequations}

MPC controls vehicle acceleration and steering to follow the reference trajectory using the dynamics model $M_{dynamics}$ in Eq.~\eqref{eq:mpc:dynamic_model}. $(x_t^{car}, y_t^{car})$ denote the vehicle’s global position; $v_t$ and $yaw_t$ are its speed ($m/s$) and yaw angle ($rad$); $a_t$ and $\theta_t^{\text{steering}}$ represent acceleration ($m/s^2$) and steering angle ($rad$). The wheelbase is $L=2.89~m$, and the time step is $dt=0.02~s$. Gaussian noise $\epsilon_1$ - $\epsilon_4$ with $\sigma=0.01$ is added to model uncertainty.


The MPC objective function is composed of five parts as shown in Eq.~\eqref{eq:mpc:obj}. The first two terms are Cross Track Error ($xte$) and Error in Heading ($eth$) computed using Eq.~\eqref{eq:xte}. and Eq.~\eqref{eq:eth} respectively. $eth$ denotes the angular disparity between the intended path direction and the current heading of a vehicle in path tracking systems. $v_t^{\text{ref}} = 10m/s$ is the desired speed of the vehicle. The last two terms regulate the rate of change of the steering angle and acceleration to make planned trajectory smoother. $w_0, w_1, w_2, w_3, w_4$ are the coefficients balancing the importance of each term. The planning horizon of the MPC is 25 steps and the planning step $dt$ is 0.02 seconds. 

\begin{equation}\label{eq:mpc:obj}
\begin{aligned}
J_t &= \sum_{i=t}^{t+H-1} ( w_0 \cdot xte_i^2 + w_1 \cdot eth_i^2 \\
&\quad+ w_2 \cdot (v_i - v_t^{\text{ref}})^2 \\
&\quad+ w_3 \cdot (steer_i - steer_{i-1})^2 \\
&\quad+ w_4 \cdot (throttle_i - throttle_{i-1})^2)
\end{aligned}
\end{equation}

\begin{align}
    xte = distance(pos^{car}_t, \mathbf{wp}_i(x_i, y_i)^{closest})
    \label{eq:xte}
\end{align}

\begin{align}
    eth = abs\left(yaw^{car}_t - \arctan \left( {\frac{y_{i+1} - y_i}{x_{i+1} - x_i}} \right) \right)
    \label{eq:eth}
\end{align}

Gradient-based solvers are highly effective when both system dynamics and cost functions are smooth and differentiable. However, in our setting, the calculation of $xte$ (cross-track error) requires finding the closest point on a reference trajectory to the vehicle’s current position, which introduces discontinuities and non-differentiability into the objective function. This motivates the use of gradient-free solvers such as COBYLA~\cite{powell1994direct}, which are better suited for non-differentiable optimization problems~\cite{lee2011model, schwenzer2021review}. Nonetheless, with high speeds and on tracks with sharp turns, gradient-free MPC solvers struggle to optimize trajectories in real time without a good initial guess, making practical deployment challenging. As shown in Section~\ref{sec:experiment_result}, the gradient-free MPC solver succeeds only on the simple IMS track but fails on all complex tracks.

We implement an early-stop condition when the planned trajectory’s accumulated $xte$ is below $0.1,m$, preventing MPC wasting time on optimizing a trajectory that is already good enough. We also cap MPC iterations at 100 during RL training to avoid solver stagnation and evaluate performance under both 100 and 200 iteration limits during testing for sensitivity analysis.


\subsubsection{RL Details}

\RALedit{The reward at each step of the RL training is shown in Eq.~\eqref{eq:rl:reward}. The first term is the negative MPC optimization time, and the second term is the negative of the $xte$ over the planning horizon, $H$. This reward design helps optimize $\pi_{\theta}^{warm}$ by minimizing the MPC running time directly while maximizing tracking accuracy.} 
\icra{
\begin{equation}\label{eq:rl:reward}
\begin{aligned}
r_t = -time_{MPC} - xte(pos_t^{car}, Traj^{ref})
\end{aligned}
\end{equation}
}

\vspace{-1em}

\subsection{Sampling Based MPC}

\subsubsection{MPC Details}

\begin{figure}[H]
    \centering
    \begin{subfigure}[b]{0.38\textwidth}
        \centering
        \includegraphics[width=\textwidth]{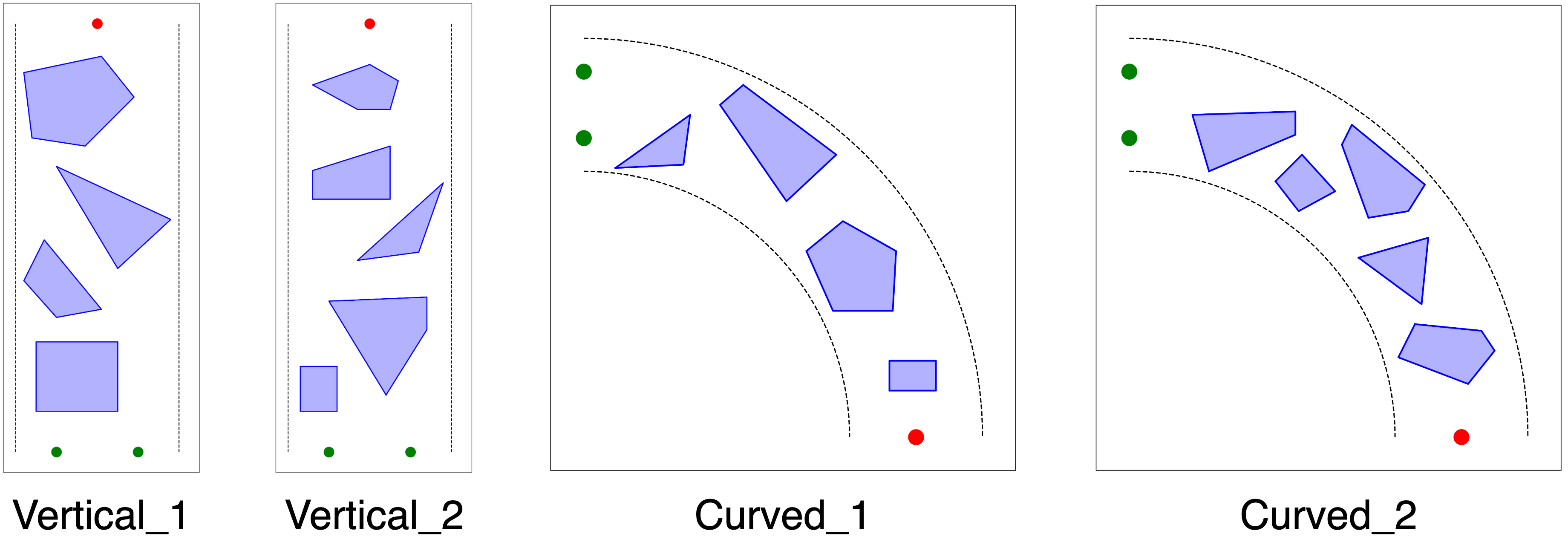}
        \caption{Training Maps}
        \label{fig:mppi_training_map}
    \end{subfigure}
    \hfill
    \begin{subfigure}[b]{0.38\textwidth}
        \centering
        \includegraphics[width=\textwidth]{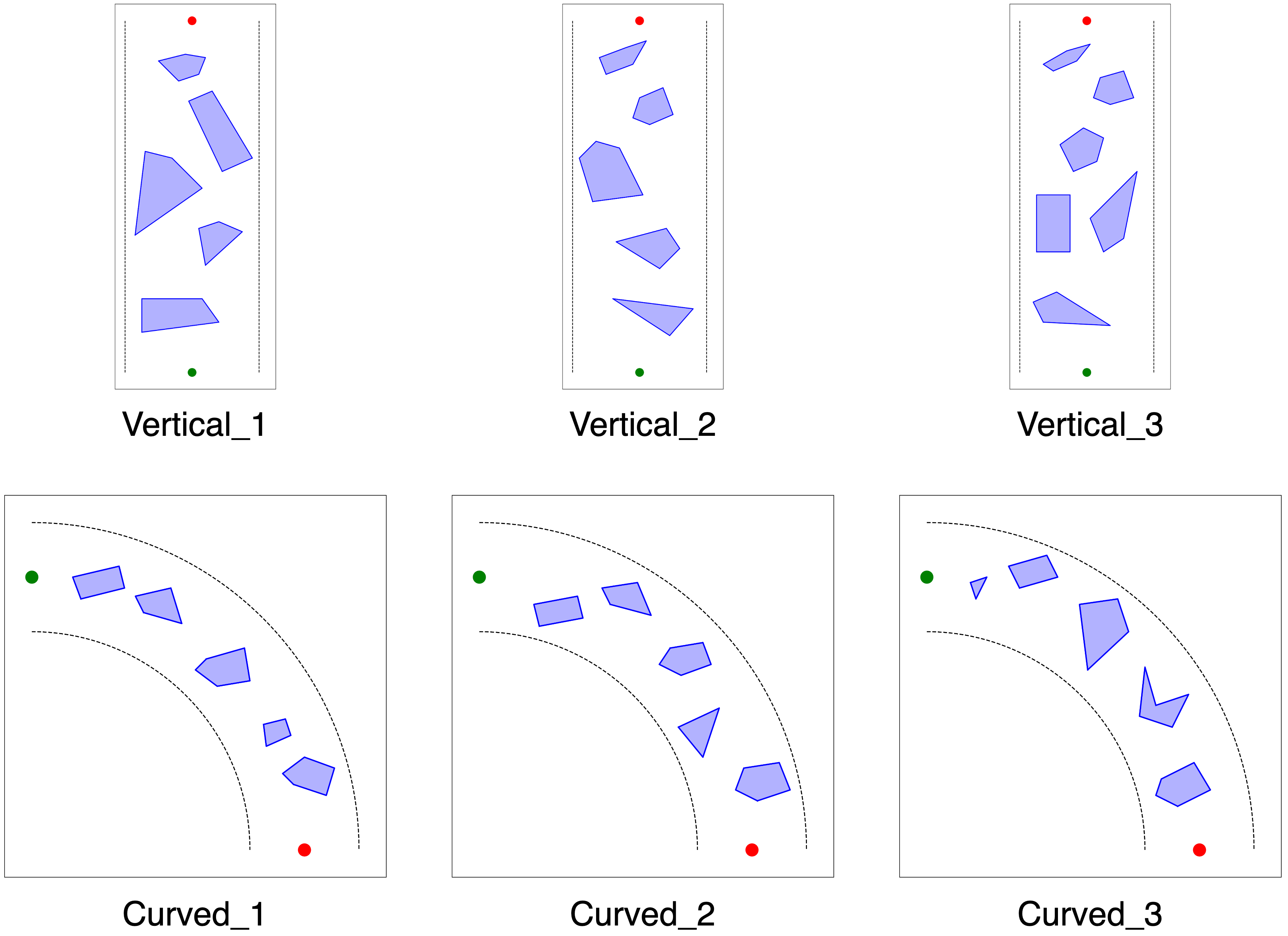}
        \caption{Testing Maps}
        \label{fig:mppi_testing_map}
    \end{subfigure}
    \caption{\RAL{Training and testing maps in the obstacle-rich navigation domain. Green dots indicate spawn points, red dots represent destinations, and polygons correspond to obstacles.}}
    \label{fig:mppi_maps}
\end{figure}

Unlike deterministic MPC, MPPI employs a sampling-based stochastic optimization framework to minimize a predefined cost. This setup allows us to test whether our warm-start algorithm can also enhance a different control optimizer. \RALedit{As MPPI is mainly used for path-planning and obstacle-avoidance tasks~\cite{zhai2025pa, streichenberg2023multi}, we build an obstacle-rich navigation domain for evaluation, as shown in Fig.~\ref{fig:mppi_maps}. The training set includes two straight and two curved-road maps, while the testing set contains three unseen vertical and three unseen curved-road maps.} Each map features randomly generated polygonal obstacles with varying sizes, shapes, and positions, posing diverse planning challenges. The vehicle starts at a green dot and must reach the goal region (red dot) without collisions. To increase training diversity, each training map offers two initialization points: one on the left and one on the right.

\begin{equation}\label{eq:mppi:obj}
\begin{aligned}
J = \sum_{i=t}^{t+H-1} &\Big[ 
w_1 \, throttle_i^2 + w_2 \, steer_i^2 \\
&\quad + w_3 \, (v_i - v_{t}^{ref})^2 \\
&\quad + w_4 \, C_{obstacle}(x_i, y_i) \\
&\quad + w_5 \, C_{border}(x_i, y_i) 
\Big] \\
& + \quad \sum_{i=t+H-3}^{t+H-1} \Big[ w_6 \, C_{goal}(x_{i}, y_{i}) \Big]
\end{aligned}
\end{equation}

The vehicle dynamics model is identical to that employed in the deterministic MPC in Eq.~\eqref{eq:mpc:dynamic_model}, ensuring a fair comparison. The cost function for MPPI is shown in Eq.~\eqref{eq:mppi:obj}. The first two terms correspond to smoothness penalties to enhance control stability. $C_{obstacle}$ penalizes proximity to obstacles, $C_{border}$ penalizes deviations beyond road boundaries to maintain lane centering, and $C_{goal}$ penalizes the final distance to the target region. The planning horizon is fixed at 25 steps. And number of samples is set to 200 samples to balance efficiency and control quality.


\subsubsection{RL Details}

The RL reward function is defined in Eq.~\eqref{eq:mppi_rl:reward}. It is aligned with the MPPI objective function. The only difference is the final term, which encourages the vehicle to reduce its distance to the goal at each timestep. \RALedit{Since MPPI operates via parallel trajectory sampling rather than iterative optimization, we exclude the MPC optimization time from the RL reward, as all trajectories are sampled simultaneously on the GPU. However, warm-starting remains valuable for onboard systems with limited parallelism. A good initial guess allows the controller to achieve comparable performance with fewer samples or higher accuracy with the same number of samples. In our evaluation, we fix the number of samples and test whether our method outperforms the baselines in control performance.}

\begin{equation}\label{eq:mppi_rl:reward}
\begin{aligned}
r_{t}^{mppi} &=  
-w_1 \, throttle_t^2 - w_2 \, steer_t^2 \\
&\quad - w_3 \, (v_t - v_{t}^{ref})^2 - w_4 \, C_{obstacle}(x_t, y_t) \\
&\quad - w_5 \, C_{border}(x_t, y_t) \\
&\quad - w_6 \, ((C_{goal}(x_{t}, y_{t}) - (C_{goal}(x_{t-1}, y_{t-1})) \\
\end{aligned}
\end{equation}

\begin{table*}[!b]
\centering
\caption{\RALedit{Deterministic MPC results with a maximum of 100 optimization iterations.} Values represent averages over three test runs with different random seeds. Reported percentages indicate relative improvement compared to the corresponding baseline. The asterisk (*) implies statistical significance (paired Wilcoxon signed-rank tests): * $p<0.05$, ** $p<0.01$, *** $p<0.001$, **** $p<0.0001$, n.s. means no significance.} \label{table:f1_results}
\resizebox{\textwidth}{!}{%
\begin{tabular}{ccccccccc}
\multicolumn{1}{c|}{} & \multicolumn{2}{c|}{\begin{tabular}[c]{@{}c@{}}All Zeros \& \\ Previous Solution \& \\ Ours w/o BC\end{tabular}} & \multicolumn{2}{c|}{\begin{tabular}[c]{@{}c@{}}Ours\\ w/o RL\end{tabular}} & \multicolumn{2}{c|}{\begin{tabular}[c]{@{}c@{}}Ours\\ w/o DAgger\end{tabular}} & \multicolumn{2}{c}{Ours} \\ \cline{2-9} 
\multicolumn{1}{c|}{} & \begin{tabular}[c]{@{}c@{}}Optimization\\ Time (Second)\end{tabular} & \multicolumn{1}{c|}{\begin{tabular}[c]{@{}c@{}}xte\\ (Meter)\end{tabular}} & \begin{tabular}[c]{@{}c@{}}Optimization\\ Time (Second)\end{tabular} & \multicolumn{1}{c|}{\begin{tabular}[c]{@{}c@{}}xte\\ (Meter)\end{tabular}} & \begin{tabular}[c]{@{}c@{}}Optimization\\ Time (Second)\end{tabular} & \multicolumn{1}{c|}{\begin{tabular}[c]{@{}c@{}}xte\\ (Meter)\end{tabular}} & \begin{tabular}[c]{@{}c@{}}Optimization\\ Time (Second)\end{tabular} & \begin{tabular}[c]{@{}c@{}}xte\\ (Meter)\end{tabular} \\ \hline
\multicolumn{9}{c}{Training Tracks} \\ \hline
\multicolumn{1}{c|}{Zandvoort} &  & \multicolumn{1}{c|}{} & 0.1270 ± 0.0800 & \multicolumn{1}{c|}{0.5929 ± 0.6066} & 0.1530 ± 0.0542 & \multicolumn{1}{c|}{\textbf{0.3869 ± 0.2840}} & \textbf{0.1087 ± 0.0821} & 0.3975 ± 0.3084 \\ \cline{1-1} \cline{4-9} 
\multicolumn{1}{c|}{Sochi} &  & \multicolumn{1}{c|}{} & 0.1429 ± 0.0999 & \multicolumn{1}{c|}{0.5657 ± 0.5869} & 0.1772 ± 0.0620 & \multicolumn{1}{c|}{0.4704 ± 0.5150} & \textbf{0.1128 ± 0.1010} & \textbf{0.3755 ± 0.3075} \\ \cline{1-1} \cline{4-9} 
\multicolumn{1}{c|}{Hockenheim} & \multicolumn{2}{c|}{Failed} & 0.1129 ± 0.0912 & \multicolumn{1}{c|}{0.4686 ± 0.4177} & 0.1450 ± 0.0581 & \multicolumn{1}{c|}{0.4078 ± 0.4665} & \textbf{0.0865 ± 0.0877} & \textbf{0.3548 ± 0.3373} \\ \cline{1-1} \cline{4-9} 
\multicolumn{1}{c|}{\begin{tabular}[c]{@{}c@{}}Training Tracks\\ Total\end{tabular}} &  & \multicolumn{1}{c|}{} & 0.1285 ± 0.0919 & \multicolumn{1}{c|}{0.5450 ± 0.5504} & 0.1597 ± 0.0600 & \multicolumn{1}{c|}{0.4246 ± 0.4385} & \textbf{0.1033 ± 0.0920} & \textbf{0.3758 ± 0.3179} \\ \cline{1-1} \cline{4-9} 
\multicolumn{1}{c|}{\begin{tabular}[c]{@{}c@{}}Compared with\\ Ours w/o RL\end{tabular}} &  & \multicolumn{1}{c|}{} &  & \multicolumn{1}{c|}{} & \textbf{\begin{tabular}[c]{@{}c@{}}-24.27\%\\ n.s.\end{tabular}} & \multicolumn{1}{c|}{\textbf{\begin{tabular}[c]{@{}c@{}}22.10\%\\ **\end{tabular}}} & \textbf{\begin{tabular}[c]{@{}c@{}}19.65\%\\ **\end{tabular}} & \textbf{\begin{tabular}[c]{@{}c@{}}30.04\%\\ **\end{tabular}} \\ \hline
\multicolumn{9}{c}{Zero-shot Tracks} \\ \hline
\multicolumn{1}{c|}{Nuerburgring} &  & \multicolumn{1}{c|}{} & 0.1845 ± 0.0669 & \multicolumn{1}{c|}{1.6630 ± 1.5488} & 0.1679 ± 0.0601 & \multicolumn{1}{c|}{\textbf{0.3774 ±0.3233}} & \textbf{0.1158 ± 0.0967} & 0.4090 ± 0.3500 \\ \cline{1-1} \cline{4-9} 
\multicolumn{1}{c|}{BrandsHatch} &  & \multicolumn{1}{c|}{} & 0.1163 ± 0.0922 & \multicolumn{1}{c|}{0.5117 ± 0.5484} & 0.1545 ± 0.0519 & \multicolumn{1}{c|}{0.3463 ± 0.2075} & \textbf{0.0895 ± 0.0837} & \textbf{0.3355 ± 0.2592} \\ \cline{1-1} \cline{4-9} 
\multicolumn{1}{c|}{Oschersleben} &  & \multicolumn{1}{c|}{} & 0.1195 ± 0.0832 & \multicolumn{1}{c|}{0.4181 ± 0.3514} & 0.1421 ± 0.0582 & \multicolumn{1}{c|}{0.4383 ± 0.4309} & \textbf{0.1116 ± 0.0846} & \textbf{0.3405 ± 0.2395} \\ \cline{1-1} \cline{4-9} 
\multicolumn{1}{c|}{Shanghai} & \multicolumn{2}{c|}{Failed} & 0.1845 ± 0.0732 & \multicolumn{1}{c|}{1.3278 ± 1.2934} & 0.1717 ± 0.0602 & \multicolumn{1}{c|}{0.4242 ± 0.4157} & \textbf{0.1189 ± 0.0977} & \textbf{0.3869 ± 0.2688} \\ \cline{1-1} \cline{4-9} 
\multicolumn{1}{c|}{Melbourne} &  & \multicolumn{1}{c|}{} & 0.1375 ± 0.0963 & \multicolumn{1}{c|}{0.7781 ± 0.9978} & 0.1767 ± 0.0588 & \multicolumn{1}{c|}{0.4297 ± 0.3378} & \textbf{0.1235 ± 0.0937} & \textbf{0.3921 ± 0.3028} \\ \cline{1-1} \cline{4-9} 
\multicolumn{1}{c|}{Catalunya} &  & \multicolumn{1}{c|}{} & 0.1140 ± 0.0942 & \multicolumn{1}{c|}{0.5561 ± 0.6710} & 0.1561 ± 0.0571 & \multicolumn{1}{c|}{0.4095 ± 0.3798} & \textbf{0.1019 ± 0.0891} & \textbf{0.3631 ± 0.2863} \\ \cline{1-1} \cline{4-9} 
\multicolumn{1}{c|}{\begin{tabular}[c]{@{}c@{}}Zero-shot Tracks\\ Total\end{tabular}} &  & \multicolumn{1}{c|}{} & 0.1558 ± 0.0871 & \multicolumn{1}{c|}{1.0696 ± 1.2447} & 0.1629 ± 0.0589 & \multicolumn{1}{c|}{0.4033 ± 0.3539} & \textbf{0.1095 ± 0.0913} & \textbf{0.3707 ± 0.2905} \\ \cline{1-1} \cline{4-9} 
\multicolumn{1}{c|}{\begin{tabular}[c]{@{}c@{}}Compared with\\ Ours w/o RL\end{tabular}} &  & \multicolumn{1}{c|}{} &  & \multicolumn{1}{c|}{} & \textbf{\begin{tabular}[c]{@{}c@{}}-4.57\%\\ n.s.\end{tabular}} & \multicolumn{1}{c|}{\textbf{\begin{tabular}[c]{@{}c@{}}62.30\%\\ ***\end{tabular}}} & \textbf{\begin{tabular}[c]{@{}c@{}}29.75\%\\ ***\end{tabular}} & \textbf{\begin{tabular}[c]{@{}c@{}}65.34\%\\ ****\end{tabular}}
\end{tabular}%
}
\end{table*}

\begin{table*}[!b]
\centering
\caption{\RALedit{Deterministic MPC results with a maximum of 200 optimization iterations.} Values represent averages over three test runs with different random seeds. Reported percentages indicate relative improvement compared to the corresponding baseline. The asterisk (*) implies statistical significance (paired Wilcoxon signed-rank tests): * $p<0.05$, ** $p<0.01$, *** $p<0.001$, **** $p<0.0001$, n.s. means no significance.} 
\label{table:f1_results_200iter}
\resizebox{\textwidth}{!}{%
\begin{tabular}{ccccccccc}
\multicolumn{1}{c|}{} & \multicolumn{2}{c|}{\begin{tabular}[c]{@{}c@{}}All Zeros \& \\ Previous Solution \& \\ Ours w/o BC\end{tabular}} & \multicolumn{2}{c|}{\begin{tabular}[c]{@{}c@{}}Ours\\ w/o RL\end{tabular}} & \multicolumn{2}{c|}{\begin{tabular}[c]{@{}c@{}}Ours\\ w/o DAgger\end{tabular}} & \multicolumn{2}{c}{Ours} \\ \cline{2-9} 
\multicolumn{1}{c|}{} & \begin{tabular}[c]{@{}c@{}}Optimization\\ Time (Second)\end{tabular} & \multicolumn{1}{c|}{\begin{tabular}[c]{@{}c@{}}xte\\ (Meter)\end{tabular}} & \begin{tabular}[c]{@{}c@{}}Optimization\\ Time (Second)\end{tabular} & \multicolumn{1}{c|}{\begin{tabular}[c]{@{}c@{}}xte\\ (Meter)\end{tabular}} & \begin{tabular}[c]{@{}c@{}}Optimization\\ Time (Second)\end{tabular} & \multicolumn{1}{c|}{\begin{tabular}[c]{@{}c@{}}xte\\ (Meter)\end{tabular}} & \begin{tabular}[c]{@{}c@{}}Optimization\\ Time (Second)\end{tabular} & \begin{tabular}[c]{@{}c@{}}xte\\ (Meter)\end{tabular} \\ \hline
\multicolumn{9}{c}{Training Tracks} \\ \hline
\multicolumn{1}{c|}{Zandvoort} &  & \multicolumn{1}{c|}{} & 0.2395 ± 0.1700 & \multicolumn{1}{c|}{0.5130 ± 0.5054} & 0.2498 ± 0.1412 & \multicolumn{1}{c|}{\textbf{0.3472 ± 0.2664}} & \textbf{0.1954 ± 0.1753} & 0.3501 ± 0.2402 \\ \cline{1-1} \cline{4-9} 
\multicolumn{1}{c|}{Sochi} &  & \multicolumn{1}{c|}{} & 0.2821 ± 0.2079 & \multicolumn{1}{c|}{0.5367 ± 0.5056} & 0.2836 ± 0.1649 & \multicolumn{1}{c|}{0.3783 ± 0.3697} & \textbf{0.2166 ± 0.2122} & \textbf{0.3621 ± 0.2745} \\ \cline{1-1} \cline{4-9} 
\multicolumn{1}{c|}{Hockenheim} & \multicolumn{2}{c|}{Failed} & 0.2072 ± 0.1914 & \multicolumn{1}{c|}{0.4245 ± 0.3663} & 0.2349 ± 0.1502 & \multicolumn{1}{c|}{0.3754 ± 0.4123} & \textbf{0.1593 ± 0.1815} & \textbf{0.3172 ± 0.2734} \\ \cline{1-1} \cline{4-9} 
\multicolumn{1}{c|}{\begin{tabular}[c]{@{}c@{}}Training Tracks\\ Total\end{tabular}} &  & \multicolumn{1}{c|}{} & 0.2451 ± 0.1935 & \multicolumn{1}{c|}{0.4944 ± 0.4695} & 0.2578 ± 0.1545 & \multicolumn{1}{c|}{0.3674 ± 0.3546} & \textbf{0.1918 ± 0.1930} & \textbf{0.3441 ± 0.2645} \\ \cline{1-1} \cline{4-9} 
\multicolumn{1}{c|}{\begin{tabular}[c]{@{}c@{}}Compared with\\ Ours w/o RL\end{tabular}} &  & \multicolumn{1}{c|}{} &  & \multicolumn{1}{c|}{} & \textbf{\begin{tabular}[c]{@{}c@{}}-5.19\%\\ n.s.\end{tabular}} & \multicolumn{1}{c|}{\textbf{\begin{tabular}[c]{@{}c@{}}25.70\%\\ **\end{tabular}}} & \textbf{\begin{tabular}[c]{@{}c@{}}21.77\%\\ **\end{tabular}} & \textbf{\begin{tabular}[c]{@{}c@{}}30.41\%\\ **\end{tabular}} \\ \hline
\multicolumn{9}{c}{Zero-shot Tracks} \\ \hline
\multicolumn{1}{c|}{Nuerburgring} &  & \multicolumn{1}{c|}{} & 0.3086 ± 0.1865 & \multicolumn{1}{c|}{0.5814 ± 0.5536} & 0.2685 ± 0.1576 & \multicolumn{1}{c|}{\textbf{0.3359 ±0.3251}} & \textbf{0.2358 ± 0.2008} & 0.4254 ± 0.3867 \\ \cline{1-1} \cline{4-9} 
\multicolumn{1}{c|}{BrandsHatch} &  & \multicolumn{1}{c|}{} & 0.2172 ± 0.1887 & \multicolumn{1}{c|}{0.4122 ± 0.3194} & 0.2514 ± 0.1421 & \multicolumn{1}{c|}{\textbf{0.3063 ± 0.1850}} & \textbf{0.1584 ± 0.1771} & 0.3213 ± 0.2282 \\ \cline{1-1} \cline{4-9} 
\multicolumn{1}{c|}{Oschersleben} &  & \multicolumn{1}{c|}{} & \textbf{0.2340 ± 0.1754} & \multicolumn{1}{c|}{0.3977 ± 0.3196} & 0.2467 ± 0.1473 & \multicolumn{1}{c|}{\textbf{0.3623 ± 0.2838}} & 0.2376 ± 0.1756 & 0.3893 ± 0.2840 \\ \cline{1-1} \cline{4-9} 
\multicolumn{1}{c|}{Shanghai} & \multicolumn{2}{c|}{Failed} & 0.3093 ± 0.1943 & \multicolumn{1}{c|}{0.6666 ± 0.6522} & 0.2806 ± 0.1551 & \multicolumn{1}{c|}{\textbf{0.3402 ± 0.3107}} & \textbf{0.2150 ± 0.2028} & 0.3765 ± 0.2549 \\ \cline{1-1} \cline{4-9} 
\multicolumn{1}{c|}{Melbourne} &  & \multicolumn{1}{c|}{} & 0.2747 ± 0.2011 & \multicolumn{1}{c|}{0.6036 ± 0.6359} & 0.2930 ± 0.1561 & \multicolumn{1}{c|}{0.3868 ± 0.3315} & \textbf{0.2223 ± 0.2016} & \textbf{0.3584 ± 0.2459} \\ \cline{1-1} \cline{4-9} 
\multicolumn{1}{c|}{Catalunya} &  & \multicolumn{1}{c|}{} & 0.2300 ± 0.1947 & \multicolumn{1}{c|}{0.5820 ± 0.6688} & 0.2663 ± 0.1521 & \multicolumn{1}{c|}{0.3896 ± 0.3693} & \textbf{0.1920 ± 0.1871} & \textbf{0.3704 ± 0.3305} \\ \cline{1-1} \cline{4-9} 
\multicolumn{1}{c|}{\begin{tabular}[c]{@{}c@{}}Zero-shot Tracks\\ Total\end{tabular}} &  & \multicolumn{1}{c|}{} & 0.2663 ± 0.1949 & \multicolumn{1}{c|}{0.5549 ± 0.5710} & 0.2697 ± 0.1531 & \multicolumn{1}{c|}{\textbf{0.3543 ± 0.3112}} & \textbf{0.2078 ± 0.1921} & 0.3689 ± 0.2909 \\ \cline{1-1} \cline{4-9} 
\multicolumn{1}{c|}{\begin{tabular}[c]{@{}c@{}}Compared with\\ Ours w/o RL\end{tabular}} &  & \multicolumn{1}{c|}{} &  & \multicolumn{1}{c|}{} & \textbf{\begin{tabular}[c]{@{}c@{}}-1.30\%\\ n.s.\end{tabular}} & \multicolumn{1}{c|}{\textbf{\begin{tabular}[c]{@{}c@{}}36.16\%\\ ***\end{tabular}}} & \textbf{\begin{tabular}[c]{@{}c@{}}21.97\%\\ ****\end{tabular}} & \textbf{\begin{tabular}[c]{@{}c@{}}33.52\%\\ ***\end{tabular}}
\end{tabular}%
}
\end{table*}

\subsection{Baselines and Metrics}

For deterministic MPC, we evaluate performance using two metrics: (1) the average optimization time per step (seconds) and (2) the average $xte$ (meters) per step. For sampling-based MPC, we employ three metrics: (1) steering standard deviation, to assess control stability; (2) average number of steps required to reach the goal, as a proxy for path efficiency; and (3) number of obstacle collisions, to evaluate safety and avoidance capability. During testing, we compare the performance of our algorithm against five types of initial-guess policies: 
\begin{itemize}
    \item \textbf{All Zero:} all-zero initial guesses. 
    \item \textbf{Previous Solution:} initial guesses derived from the MPC solution at the previous step.
    \item \textbf{Ours w/o BC:} initial guesses from a warm-start policy trained exclusively with an online training algorithm, without offline training. 
    \item \textbf{Ours w/o RL:} initial guesses from a warm-start policy trained solely with offline BC. 
    \item \textbf{Ours w/o DAgger:} initial guesses from a warm-start policy trained with our two-phase learning algorithm but without the use of DAgger. 
\end{itemize}
To ensure a fair comparison between the Ours w/o BC and warm-start policy trained by our two-phase learning algorithm, we extend the training time of Ours w/o BC by an additional hour to account for the data collection time in the offline BC phase.


\begin{table*}[!b]
\centering
\caption{\RALedit{Sampling Based MPC Results. Values represent averages over three test runs with different random seeds. Numbers in parentheses denote the standard error.}} 
\label{table:mppi_results}
\resizebox{\textwidth}{!}{%
\begin{tabular}{ccccccccccccccccccc}
 & \multicolumn{3}{c|}{All Zeros} & \multicolumn{3}{c|}{Previous Solution} & \multicolumn{3}{c|}{\begin{tabular}[c]{@{}c@{}}Ours\\ w/o BC\end{tabular}} & \multicolumn{3}{c|}{\begin{tabular}[c]{@{}c@{}}Ours\\ w/o RL\end{tabular}} & \multicolumn{3}{c|}{\begin{tabular}[c]{@{}c@{}}Ours\\ w/o DAgger\end{tabular}} & \multicolumn{3}{c}{Ours} \\ \hline
\multicolumn{1}{c|}{\begin{tabular}[c]{@{}c@{}}Map\\ Name\end{tabular}} & \begin{tabular}[c]{@{}c@{}}Steer\\ Std\end{tabular} & \begin{tabular}[c]{@{}c@{}}Number\\ of Steps\end{tabular} & \multicolumn{1}{c|}{\begin{tabular}[c]{@{}c@{}}Number\\ of Hits\end{tabular}} & \begin{tabular}[c]{@{}c@{}}Steer\\ Std\end{tabular} & \begin{tabular}[c]{@{}c@{}}Number\\ of Steps\end{tabular} & \multicolumn{1}{c|}{\begin{tabular}[c]{@{}c@{}}Number\\ of Hits\end{tabular}} & \begin{tabular}[c]{@{}c@{}}Steer\\ Std\end{tabular} & \begin{tabular}[c]{@{}c@{}}Number\\ of Steps\end{tabular} & \multicolumn{1}{c|}{\begin{tabular}[c]{@{}c@{}}Number\\ of Hits\end{tabular}} & \begin{tabular}[c]{@{}c@{}}Steer\\ Std\end{tabular} & \begin{tabular}[c]{@{}c@{}}Number\\ of Steps\end{tabular} & \multicolumn{1}{c|}{\begin{tabular}[c]{@{}c@{}}Number\\ of Hits\end{tabular}} & \begin{tabular}[c]{@{}c@{}}Steer\\ Std\end{tabular} & \begin{tabular}[c]{@{}c@{}}Number\\ of Steps\end{tabular} & \multicolumn{1}{c|}{\begin{tabular}[c]{@{}c@{}}Number\\ of Hits\end{tabular}} & \begin{tabular}[c]{@{}c@{}}Steer\\ Std\end{tabular} & \begin{tabular}[c]{@{}c@{}}Number\\ of Steps\end{tabular} & \begin{tabular}[c]{@{}c@{}}Number\\ of Hits\end{tabular} \\ \hline
\multicolumn{19}{c}{Training Maps} \\ \hline
\multicolumn{1}{c|}{Vertical\_1} & 0.6647 & \begin{tabular}[c]{@{}c@{}}287.3333\\ ± 22.8356\end{tabular} & \multicolumn{1}{c|}{\begin{tabular}[c]{@{}c@{}}2.0000\\ ± 1.4142\end{tabular}} & 0.6306 & \begin{tabular}[c]{@{}c@{}}251.8333\\ ± 12.0568\end{tabular} & \multicolumn{1}{c|}{\textbf{\begin{tabular}[c]{@{}c@{}}0.0000\\ ± 0.0000\end{tabular}}} & 0.4881 & \begin{tabular}[c]{@{}c@{}}258.0000\\ ± 19.7281\end{tabular} & \multicolumn{1}{c|}{\begin{tabular}[c]{@{}c@{}}1.0000\\ ± 1.0954\end{tabular}} & 0.4747 & \begin{tabular}[c]{@{}c@{}}278.1667\\ ± 16.7978\end{tabular} & \multicolumn{1}{c|}{\begin{tabular}[c]{@{}c@{}}0.6667\\ ± 0.8165\end{tabular}} & 0.4712 & \textbf{\begin{tabular}[c]{@{}c@{}}245.1667\\ ± 20.8750\end{tabular}} & \multicolumn{1}{c|}{\begin{tabular}[c]{@{}c@{}}0.8333\\ ± 1.6021\end{tabular}} & \textbf{0.4689} & \begin{tabular}[c]{@{}c@{}}249.3333\\ ± 22.2860\end{tabular} & \textbf{\begin{tabular}[c]{@{}c@{}}0.0000\\ ± 0.0000\end{tabular}} \\ \hline
\multicolumn{1}{c|}{Vertical\_2} & 0.6491 & \begin{tabular}[c]{@{}c@{}}289.0000\\ ± 45.4093\end{tabular} & \multicolumn{1}{c|}{\begin{tabular}[c]{@{}c@{}}0.6667\\ ± 1.2111\end{tabular}} & 0.6393 & \begin{tabular}[c]{@{}c@{}}273.6667\\ ± 11.5873\end{tabular} & \multicolumn{1}{c|}{\textbf{\begin{tabular}[c]{@{}c@{}}0.0000\\ ± 0.0000\end{tabular}}} & 0.4923 & \begin{tabular}[c]{@{}c@{}}264.0000\\ ± 23.0304\end{tabular} & \multicolumn{1}{c|}{\begin{tabular}[c]{@{}c@{}}0.8333\\ ± 0.9832\end{tabular}} & \textbf{0.4637} & \begin{tabular}[c]{@{}c@{}}279.8333\\ ± 24.7258\end{tabular} & \multicolumn{1}{c|}{\textbf{\begin{tabular}[c]{@{}c@{}}0.0000\\ ± 0.0000\end{tabular}}} & 0.4810 & \begin{tabular}[c]{@{}c@{}}265.6667\\ ± 19.2942\end{tabular} & \multicolumn{1}{c|}{\textbf{\begin{tabular}[c]{@{}c@{}}0.0000\\ ± 0.0000\end{tabular}}} & 0.4716 & \textbf{\begin{tabular}[c]{@{}c@{}}256.5000\\ ± 26.8384\end{tabular}} & \begin{tabular}[c]{@{}c@{}}0.1667\\ ± 0.4082\end{tabular} \\ \hline
\multicolumn{1}{c|}{Curved\_1} & 0.6290 & \begin{tabular}[c]{@{}c@{}}369.8333\\ ± 37.0751\end{tabular} & \multicolumn{1}{c|}{\begin{tabular}[c]{@{}c@{}}1.5000\\ ± 3.6742\end{tabular}} & 0.6009 & \begin{tabular}[c]{@{}c@{}}360.3333\\ ± 13.4263\end{tabular} & \multicolumn{1}{c|}{\textbf{\begin{tabular}[c]{@{}c@{}}0.0000\\ ± 0.0000\end{tabular}}} & 0.4773 & \begin{tabular}[c]{@{}c@{}}376.3333\\ ± 84.5167\end{tabular} & \multicolumn{1}{c|}{\textbf{\begin{tabular}[c]{@{}c@{}}0.0000\\ ± 0.0000\end{tabular}}} & 0.4748 & \begin{tabular}[c]{@{}c@{}}354.6667\\ ± 36.9035\end{tabular} & \multicolumn{1}{c|}{\textbf{\begin{tabular}[c]{@{}c@{}}0.0000\\ ± 0.0000\end{tabular}}} & 0.4322 & \textbf{\begin{tabular}[c]{@{}c@{}}334.6667\\ ± 4.1793\end{tabular}} & \multicolumn{1}{c|}{\begin{tabular}[c]{@{}c@{}}0.0000\\ ± 0.0000\end{tabular}} & \textbf{0.4256} & \begin{tabular}[c]{@{}c@{}}335.0000\\ ± 5.6569\end{tabular} & \textbf{\begin{tabular}[c]{@{}c@{}}0.0000\\ ± 0.0000\end{tabular}} \\ \hline
\multicolumn{1}{c|}{Curved\_2} & 0.6366 & \begin{tabular}[c]{@{}c@{}}422.6667\\ ± 65.9019\end{tabular} & \multicolumn{1}{c|}{\begin{tabular}[c]{@{}c@{}}9.3333\\ ± 17.5803\end{tabular}} & 0.6369 & \begin{tabular}[c]{@{}c@{}}401.3333\\ ± 26.1738\end{tabular} & \multicolumn{1}{c|}{\textbf{\begin{tabular}[c]{@{}c@{}}0.0000\\ ± 0.0000\end{tabular}}} & 0.5064 & \begin{tabular}[c]{@{}c@{}}427.6667\\ ± 60.8857\end{tabular} & \multicolumn{1}{c|}{\begin{tabular}[c]{@{}c@{}}1.0000\\ ± 2.4495\end{tabular}} & 0.4879 & \begin{tabular}[c]{@{}c@{}}460.5000\\ ± 81.4389\end{tabular} & \multicolumn{1}{c|}{\textbf{\begin{tabular}[c]{@{}c@{}}0.0000\\ ± 0.0000\end{tabular}}} & 0.4812 & \textbf{\begin{tabular}[c]{@{}c@{}}375.6667\\ ± 24.7117\end{tabular}} & \multicolumn{1}{c|}{\begin{tabular}[c]{@{}c@{}}1.3333\\ ± 2.0656\end{tabular}} & \textbf{0.4800} & \begin{tabular}[c]{@{}c@{}}376.1667\\ ± 17.6682\end{tabular} & \textbf{\begin{tabular}[c]{@{}c@{}}0.0000\\ ± 0.0000\end{tabular}} \\ \hline
\multicolumn{1}{c|}{\begin{tabular}[c]{@{}c@{}}Training Map\\ Total\end{tabular}} & 0.6433 & \begin{tabular}[c]{@{}c@{}}342.2083\\ ± 72.2255\end{tabular} & \multicolumn{1}{c|}{\begin{tabular}[c]{@{}c@{}}3.3750\\ ± 9.1358\end{tabular}} & 0.6266 & \begin{tabular}[c]{@{}c@{}}321.7917\\ ± 64.5587\end{tabular} & \multicolumn{1}{c|}{\textbf{\begin{tabular}[c]{@{}c@{}}0.0000\\ ± 0.0000\end{tabular}}} & 0.4922 & \begin{tabular}[c]{@{}c@{}}331.5000\\ ± 89.9633\end{tabular} & \multicolumn{1}{c|}{\begin{tabular}[c]{@{}c@{}}0.7083\\ ± 1.3981\end{tabular}} & 0.4775 & \begin{tabular}[c]{@{}c@{}}343.2917\\ ± 87.7870\end{tabular} & \multicolumn{1}{c|}{\begin{tabular}[c]{@{}c@{}}0.1667\\ ± 0.4815\end{tabular}} & 0.4663 & \begin{tabular}[c]{@{}c@{}}305.2917\\ ± 56.4088\end{tabular} & \multicolumn{1}{c|}{\begin{tabular}[c]{@{}c@{}}0.5417\\ ± 1.3507\end{tabular}} & \textbf{0.4619} & \textbf{\begin{tabular}[c]{@{}c@{}}304.2500\\ ± 57.5917\end{tabular}} & \begin{tabular}[c]{@{}c@{}}0.0417\\ ± 0.2041\end{tabular} \\ \hline
\multicolumn{19}{c}{Testing Maps} \\ \hline
\multicolumn{1}{c|}{Vertical\_1} & 0.6425 & \begin{tabular}[c]{@{}c@{}}285.6667\\ ± 107.7698\end{tabular} & \multicolumn{1}{c|}{\begin{tabular}[c]{@{}c@{}}9.6667\\ ± 7.7675\end{tabular}} & 0.6228 & \begin{tabular}[c]{@{}c@{}}288.3333\\ ± 97.6183\end{tabular} & \multicolumn{1}{c|}{\textbf{\begin{tabular}[c]{@{}c@{}}0.0000\\ ± 0.0000\end{tabular}}} & 0.4667 & \begin{tabular}[c]{@{}c@{}}225.0000\\ ± 1.0000\end{tabular} & \multicolumn{1}{c|}{\textbf{\begin{tabular}[c]{@{}c@{}}0.0000\\ ± 0.0000\end{tabular}}} & 0.4982 & \begin{tabular}[c]{@{}c@{}}355.6667\\ ± 198.4574\end{tabular} & \multicolumn{1}{c|}{\textbf{\begin{tabular}[c]{@{}c@{}}0.0000\\ ± 0.0000\end{tabular}}} & 0.4697 & \begin{tabular}[c]{@{}c@{}}230.0000\\ ± 21.7945\end{tabular} & \multicolumn{1}{c|}{\begin{tabular}[c]{@{}c@{}}0.3333\\ ± 0.5774\end{tabular}} & \textbf{0.4579} & \textbf{\begin{tabular}[c]{@{}c@{}}220.6667\\ ± 4.5093\end{tabular}} & \textbf{\begin{tabular}[c]{@{}c@{}}0.0000\\ ± 0.0000\end{tabular}} \\ \hline
\multicolumn{1}{c|}{Vertical\_2} & 0.6416 & \begin{tabular}[c]{@{}c@{}}266.3333\\ ± 35.5575\end{tabular} & \multicolumn{1}{c|}{\begin{tabular}[c]{@{}c@{}}2.0000\\ ± 2.6458\end{tabular}} & 0.6373 & \begin{tabular}[c]{@{}c@{}}231.6667\\ ± 3.0551\end{tabular} & \multicolumn{1}{c|}{\textbf{\begin{tabular}[c]{@{}c@{}}0.0000\\ ± 0.0000\end{tabular}}} & 0.5068 & \begin{tabular}[c]{@{}c@{}}235.6667\\ ± 14.2244\end{tabular} & \multicolumn{1}{c|}{\begin{tabular}[c]{@{}c@{}}1.0000\\ ± 1.0000\end{tabular}} & 0.4709 & \begin{tabular}[c]{@{}c@{}}225.3333\\ ± 4.1633\end{tabular} & \multicolumn{1}{c|}{\begin{tabular}[c]{@{}c@{}}0.6667\\ ± 1.1547\end{tabular}} & 0.4592 & \textbf{\begin{tabular}[c]{@{}c@{}}217.6667\\ ± 3.2146\end{tabular}} & \multicolumn{1}{c|}{\textbf{\begin{tabular}[c]{@{}c@{}}0.0000\\ ± 0.0000\end{tabular}}} & \textbf{0.4421} & \begin{tabular}[c]{@{}c@{}}219.3333\\ ± 14.4338\end{tabular} & \textbf{\begin{tabular}[c]{@{}c@{}}0.0000\\ ± 0.0000\end{tabular}} \\ \hline
\multicolumn{1}{c|}{Vertical\_3} & 0.6298 & \begin{tabular}[c]{@{}c@{}}312.0000\\ ± 106.2262\end{tabular} & \multicolumn{1}{c|}{\begin{tabular}[c]{@{}c@{}}3.3333\\ ± 4.9329\end{tabular}} & 0.6280 & \begin{tabular}[c]{@{}c@{}}263.6667\\ ± 38.5011\end{tabular} & \multicolumn{1}{c|}{\textbf{\begin{tabular}[c]{@{}c@{}}0.0000\\ ± 0.0000\end{tabular}}} & \textbf{0.4332} & \textbf{\begin{tabular}[c]{@{}c@{}}220.3333\\ ± 1.1547\end{tabular}} & \multicolumn{1}{c|}{\textbf{\begin{tabular}[c]{@{}c@{}}0.0000\\ ± 0.0000\end{tabular}}} & 0.4646 & \begin{tabular}[c]{@{}c@{}}258.0000\\ ± 4.3589\end{tabular} & \multicolumn{1}{c|}{\textbf{\begin{tabular}[c]{@{}c@{}}0.0000\\ ± 0.0000\end{tabular}}} & 0.4685 & \begin{tabular}[c]{@{}c@{}}257.3333\\ ± 1.5275\end{tabular} & \multicolumn{1}{c|}{\textbf{\begin{tabular}[c]{@{}c@{}}0.0000\\ ± 0.0000\end{tabular}}} & 0.4621 & \begin{tabular}[c]{@{}c@{}}239.0000\\ ± 25.1197\end{tabular} & \textbf{\begin{tabular}[c]{@{}c@{}}0.0000\\ ± 0.0000\end{tabular}} \\ \hline
\multicolumn{1}{c|}{Curved\_1} & 0.6251 & \begin{tabular}[c]{@{}c@{}}348.0000\\ ± 2.6458\end{tabular} & \multicolumn{1}{c|}{\begin{tabular}[c]{@{}c@{}}9.6667\\ ± 12.6623\end{tabular}} & 0.5929 & \begin{tabular}[c]{@{}c@{}}328.6667\\ ± 3.0551\end{tabular} & \multicolumn{1}{c|}{\textbf{\begin{tabular}[c]{@{}c@{}}0.0000\\ ± 0.0000\end{tabular}}} & 0.4457 & \begin{tabular}[c]{@{}c@{}}323.6667\\ ± 3.5119\end{tabular} & \multicolumn{1}{c|}{\textbf{\begin{tabular}[c]{@{}c@{}}0.0000\\ ± 0.0000\end{tabular}}} & 0.4385 & \begin{tabular}[c]{@{}c@{}}328.0000\\ ± 10.5357\end{tabular} & \multicolumn{1}{c|}{\textbf{\begin{tabular}[c]{@{}c@{}}0.0000\\ ± 0.0000\end{tabular}}} & 0.4267 & \begin{tabular}[c]{@{}c@{}}317.6667\\ ± 0.5774\end{tabular} & \multicolumn{1}{c|}{\textbf{\begin{tabular}[c]{@{}c@{}}0.0000\\ ± 0.0000\end{tabular}}} & \textbf{0.4227} & \textbf{\begin{tabular}[c]{@{}c@{}}315.6667\\ ± 1.1547\end{tabular}} & \textbf{\begin{tabular}[c]{@{}c@{}}0.0000\\ ± 0.0000\end{tabular}} \\ \hline
\multicolumn{1}{c|}{Curved\_2} & 0.5890 & \begin{tabular}[c]{@{}c@{}}331.0000\\ ± 9.1652\end{tabular} & \multicolumn{1}{c|}{\begin{tabular}[c]{@{}c@{}}0.3333\\ ± 0.5774\end{tabular}} & 0.5720 & \begin{tabular}[c]{@{}c@{}}319.6667\\ ± 0.5774\end{tabular} & \multicolumn{1}{c|}{\textbf{\begin{tabular}[c]{@{}c@{}}0.0000\\ ± 0.0000\end{tabular}}} & 0.4163 & \begin{tabular}[c]{@{}c@{}}314.0000\\ ± 1.7321\end{tabular} & \multicolumn{1}{c|}{\textbf{\begin{tabular}[c]{@{}c@{}}0.0000\\ ± 0.0000\end{tabular}}} & \textbf{0.4060} & \begin{tabular}[c]{@{}c@{}}310.6667\\ ± 1.1547\end{tabular} & \multicolumn{1}{c|}{\textbf{\begin{tabular}[c]{@{}c@{}}0.0000\\ ± 0.0000\end{tabular}}} & 0.4130 & \textbf{\begin{tabular}[c]{@{}c@{}}308.6667\\ ± 2.0817\end{tabular}} & \multicolumn{1}{c|}{\textbf{\begin{tabular}[c]{@{}c@{}}0.0000\\ ± 0.0000\end{tabular}}} & 0.4120 & \begin{tabular}[c]{@{}c@{}}310.0000\\ ± 4.5826\end{tabular} & \textbf{\begin{tabular}[c]{@{}c@{}}0.0000\\ ± 0.0000\end{tabular}} \\ \hline
\multicolumn{1}{c|}{Curved\_3} & 0.6416 & \begin{tabular}[c]{@{}c@{}}418.0000\\ ± 83.5165\end{tabular} & \multicolumn{1}{c|}{\begin{tabular}[c]{@{}c@{}}3.3333\\ ± 3.5119\end{tabular}} & 0.5841 & \begin{tabular}[c]{@{}c@{}}339.3333\\ ± 8.1445\end{tabular} & \multicolumn{1}{c|}{\textbf{\begin{tabular}[c]{@{}c@{}}0.0000\\ ± 0.0000\end{tabular}}} & 0.4654 & \begin{tabular}[c]{@{}c@{}}340.6667\\ ± 26.3881\end{tabular} & \multicolumn{1}{c|}{\textbf{\begin{tabular}[c]{@{}c@{}}0.0000\\ ± 0.0000\end{tabular}}} & 0.4975 & \begin{tabular}[c]{@{}c@{}}387.6667\\ ± 69.8164\end{tabular} & \multicolumn{1}{c|}{\begin{tabular}[c]{@{}c@{}}1.3333\\ ± 2.3094\end{tabular}} & 0.4446 & \textbf{\begin{tabular}[c]{@{}c@{}}324.3333\\ ± 5.6862\end{tabular}} & \multicolumn{1}{c|}{\textbf{\begin{tabular}[c]{@{}c@{}}0.0000\\ ± 0.0000\end{tabular}}} & \textbf{0.4330} & \begin{tabular}[c]{@{}c@{}}326.0000\\ ± 7.2111\end{tabular} & \textbf{\begin{tabular}[c]{@{}c@{}}0.0000\\ ± 0.0000\end{tabular}} \\ \hline
\multicolumn{1}{c|}{\begin{tabular}[c]{@{}c@{}}Testing Map\\ Total\end{tabular}} & 0.6284 & \begin{tabular}[c]{@{}c@{}}326.8333\\ ± 78.7821\end{tabular} & \multicolumn{1}{c|}{\begin{tabular}[c]{@{}c@{}}4.7222\\ ± 6.7196\end{tabular}} & 0.6050 & \begin{tabular}[c]{@{}c@{}}295.2222\\ ± 53.3728\end{tabular} & \multicolumn{1}{c|}{\textbf{\begin{tabular}[c]{@{}c@{}}0.0000\\ ± 0.0000\end{tabular}}} & 0.4550 & \begin{tabular}[c]{@{}c@{}}276.5556\\ ± 52.8597\end{tabular} & \multicolumn{1}{c|}{\begin{tabular}[c]{@{}c@{}}0.1667\\ ± 0.5145\end{tabular}} & 0.4713 & \begin{tabular}[c]{@{}c@{}}310.8889\\ ± 91.9545\end{tabular} & \multicolumn{1}{c|}{\begin{tabular}[c]{@{}c@{}}0.3333\\ ± 1.0290\end{tabular}} & 0.4467 & \begin{tabular}[c]{@{}c@{}}275.9444\\ ± 44.7667\end{tabular} & \multicolumn{1}{c|}{\begin{tabular}[c]{@{}c@{}}0.0556\\ ± 0.2357\end{tabular}} & \textbf{0.4372} & \textbf{\begin{tabular}[c]{@{}c@{}}271.7778\\ ± 48.6047\end{tabular}} & \textbf{\begin{tabular}[c]{@{}c@{}}0.0000\\ ± 0.0000\end{tabular}} \\ \hline
\end{tabular}%
}
\end{table*}

\begin{table*}[!b]
\centering
\caption{\RALedit{Relative improvement between different methods in sampling-based MPC and statistical test results. The reported percentages indicate relative improvement. The asterisk (*) implies statistical significance (paired Wilcoxon signed-rank tests): * $p<0.05$, ** $p<0.01$, *** $p<0.001$, **** $p<0.0001$, n.s. means no significance.}} 
\label{table:mppi_results_ablation}
\resizebox{\textwidth}{!}{%
\begin{tabular}{cccccccccccccccc}
\multicolumn{1}{c|}{\multirow{2}{*}{}} & \multicolumn{3}{c|}{Previous Solution} & \multicolumn{3}{c|}{\begin{tabular}[c]{@{}c@{}}Ours\\ w/o BC\end{tabular}} & \multicolumn{3}{c|}{\begin{tabular}[c]{@{}c@{}}Ours\\ w/o RL\end{tabular}} & \multicolumn{3}{c|}{\begin{tabular}[c]{@{}c@{}}Ours\\ w/o DAgger\end{tabular}} & \multicolumn{3}{c}{Ours} \\ \cline{2-16} 
\multicolumn{1}{c|}{} & \begin{tabular}[c]{@{}c@{}}Steer\\ Std\end{tabular} & \begin{tabular}[c]{@{}c@{}}Number\\ of Steps\end{tabular} & \multicolumn{1}{c|}{\begin{tabular}[c]{@{}c@{}}Number\\ of Hits\end{tabular}} & \begin{tabular}[c]{@{}c@{}}Steer\\ Std\end{tabular} & \begin{tabular}[c]{@{}c@{}}Number\\ of Steps\end{tabular} & \multicolumn{1}{c|}{\begin{tabular}[c]{@{}c@{}}Number\\ of Hits\end{tabular}} & \begin{tabular}[c]{@{}c@{}}Steer\\ Std\end{tabular} & \begin{tabular}[c]{@{}c@{}}Number\\ of Steps\end{tabular} & \multicolumn{1}{c|}{\begin{tabular}[c]{@{}c@{}}Number\\ of Hits\end{tabular}} & \begin{tabular}[c]{@{}c@{}}Steer\\ Std\end{tabular} & \begin{tabular}[c]{@{}c@{}}Number\\ of Steps\end{tabular} & \multicolumn{1}{c|}{\begin{tabular}[c]{@{}c@{}}Number\\ of Hits\end{tabular}} & \begin{tabular}[c]{@{}c@{}}Steer\\ Std\end{tabular} & \begin{tabular}[c]{@{}c@{}}Number\\ of Steps\end{tabular} & \begin{tabular}[c]{@{}c@{}}Number\\ of Hits\end{tabular} \\ \hline
\multicolumn{16}{c}{Training Maps} \\ \hline
\multicolumn{1}{c|}{\begin{tabular}[c]{@{}c@{}}Compared with\\ All Zeros\end{tabular}} & \textbf{\begin{tabular}[c]{@{}c@{}}2.60\%\\ **\end{tabular}} & \textbf{\begin{tabular}[c]{@{}c@{}}5.97\%\\ *\end{tabular}} & \multicolumn{1}{c|}{\textbf{\begin{tabular}[c]{@{}c@{}}100\%\\ **\end{tabular}}} & \textbf{\begin{tabular}[c]{@{}c@{}}23.49\%\\ ****\end{tabular}} & \textbf{\begin{tabular}[c]{@{}c@{}}3.13\%\\ n.s.\end{tabular}} & \multicolumn{1}{c|}{\textbf{\begin{tabular}[c]{@{}c@{}}79.01\%\\ *\end{tabular}}} & \textbf{\begin{tabular}[c]{@{}c@{}}25.77\%\\ ****\end{tabular}} & \textbf{\begin{tabular}[c]{@{}c@{}}-0.32\%\\ n.s.\end{tabular}} & \multicolumn{1}{c|}{\textbf{\begin{tabular}[c]{@{}c@{}}95.06\%\\ **\end{tabular}}} & \textbf{\begin{tabular}[c]{@{}c@{}}27.51\%\\ ****\end{tabular}} & \textbf{\begin{tabular}[c]{@{}c@{}}10.79\%\\ ****\end{tabular}} & \multicolumn{1}{c|}{\textbf{\begin{tabular}[c]{@{}c@{}}83.95\%\\ *\end{tabular}}} & \textbf{\begin{tabular}[c]{@{}c@{}}28.20\%\\ ****\end{tabular}} & \textbf{\begin{tabular}[c]{@{}c@{}}11.09\%\\ ****\end{tabular}} & \textbf{\begin{tabular}[c]{@{}c@{}}98.76\%\\ **\end{tabular}} \\ \hline
\multicolumn{1}{c|}{\begin{tabular}[c]{@{}c@{}}Compared with\\ Previous Solution\end{tabular}} & \textbf{} & \textbf{} & \multicolumn{1}{c|}{\textbf{}} & \textbf{\begin{tabular}[c]{@{}c@{}}21.45\%\\ ****\end{tabular}} & \textbf{\begin{tabular}[c]{@{}c@{}}-3.02\%\\ n.s.\end{tabular}} & \multicolumn{1}{c|}{\textbf{-}} & \textbf{\begin{tabular}[c]{@{}c@{}}23.80\%\\ ****\end{tabular}} & \textbf{\begin{tabular}[c]{@{}c@{}}-6.68\%\\ n.s.\end{tabular}} & \multicolumn{1}{c|}{\textbf{-}} & \textbf{\begin{tabular}[c]{@{}c@{}}25.58\%\\ ****\end{tabular}} & \textbf{\begin{tabular}[c]{@{}c@{}}5.13\%\\ **\end{tabular}} & \multicolumn{1}{c|}{\textbf{-}} & \textbf{\begin{tabular}[c]{@{}c@{}}26.27\%\\ ****\end{tabular}} & \textbf{\begin{tabular}[c]{@{}c@{}}5.46\%\\ **\end{tabular}} & \textbf{-} \\ \hline
\multicolumn{1}{c|}{\begin{tabular}[c]{@{}c@{}}Compared with\\ Ours w/o BC\end{tabular}} & \textbf{} & \textbf{} & \multicolumn{1}{c|}{\textbf{}} & \textbf{} & \textbf{} & \multicolumn{1}{c|}{\textbf{}} & \textbf{\begin{tabular}[c]{@{}c@{}}2.99\%\\ **\end{tabular}} & \textbf{\begin{tabular}[c]{@{}c@{}}-3.56\%\\ n.s.\end{tabular}} & \multicolumn{1}{c|}{\textbf{\begin{tabular}[c]{@{}c@{}}76.46\%\\ *\end{tabular}}} & \textbf{\begin{tabular}[c]{@{}c@{}}5.26\%\\ **\end{tabular}} & \textbf{\begin{tabular}[c]{@{}c@{}}7.91\%\\ **\end{tabular}} & \multicolumn{1}{c|}{\textbf{\begin{tabular}[c]{@{}c@{}}23.52\%\\ n.s.\end{tabular}}} & \textbf{\begin{tabular}[c]{@{}c@{}}6.16\%\\ ***\end{tabular}} & \textbf{\begin{tabular}[c]{@{}c@{}}8.22\%\\ ***\end{tabular}} & \textbf{\begin{tabular}[c]{@{}c@{}}94.11\%\\ **\end{tabular}} \\ \hline
\multicolumn{1}{c|}{\begin{tabular}[c]{@{}c@{}}Compared with\\ Ours w/o RL\end{tabular}} &  &  & \multicolumn{1}{c|}{} &  &  & \multicolumn{1}{c|}{} &  &  & \multicolumn{1}{c|}{} & \textbf{\begin{tabular}[c]{@{}c@{}}2.35\%\\ n.s.\end{tabular}} & \textbf{\begin{tabular}[c]{@{}c@{}}11.07\%\\ ***\end{tabular}} & \multicolumn{1}{c|}{\textbf{\begin{tabular}[c]{@{}c@{}}-224.96\%\\ n.s.\end{tabular}}} & \textbf{\begin{tabular}[c]{@{}c@{}}3.27\%\\ n.s.\end{tabular}} & \textbf{\begin{tabular}[c]{@{}c@{}}11.37\%\\ **\end{tabular}} & \textbf{\begin{tabular}[c]{@{}c@{}}74.99\%\\ n.s.\end{tabular}} \\ \hline
\multicolumn{1}{c|}{\begin{tabular}[c]{@{}c@{}}Compared with\\ Ours w/o DAgger\end{tabular}} &  &  & \multicolumn{1}{c|}{} &  &  & \multicolumn{1}{c|}{} &  &  & \multicolumn{1}{c|}{} & \textbf{} & \textbf{} & \multicolumn{1}{c|}{\textbf{}} & \textbf{\begin{tabular}[c]{@{}c@{}}0.95\%\\ n.s.\end{tabular}} & \textbf{\begin{tabular}[c]{@{}c@{}}0.34\%\\ n.s.\end{tabular}} & \textbf{\begin{tabular}[c]{@{}c@{}}92.30\%\\ *\end{tabular}} \\ \hline
\multicolumn{16}{c}{Testing Maps} \\ \hline
\multicolumn{1}{c|}{\begin{tabular}[c]{@{}c@{}}Compared with\\ All Zeros\end{tabular}} & \textbf{\begin{tabular}[c]{@{}c@{}}3.72\%\\ **\end{tabular}} & \textbf{\begin{tabular}[c]{@{}c@{}}9.67\%\\ **\end{tabular}} & \multicolumn{1}{c|}{\textbf{\begin{tabular}[c]{@{}c@{}}100.00\%\\ **\end{tabular}}} & \textbf{\begin{tabular}[c]{@{}c@{}}27.59\%\\ ****\end{tabular}} & \textbf{\begin{tabular}[c]{@{}c@{}}15.38\%\\ **\end{tabular}} & \multicolumn{1}{c|}{\textbf{\begin{tabular}[c]{@{}c@{}}96.47\%\\ **\end{tabular}}} & \textbf{\begin{tabular}[c]{@{}c@{}}25.00\%\\ ****\end{tabular}} & \textbf{\begin{tabular}[c]{@{}c@{}}4.88\%\\ *\end{tabular}} & \multicolumn{1}{c|}{\textbf{\begin{tabular}[c]{@{}c@{}}92.94\%\\ **\end{tabular}}} & \textbf{\begin{tabular}[c]{@{}c@{}}28.91\%\\ ****\end{tabular}} & \textbf{\begin{tabular}[c]{@{}c@{}}15.57\%\\ ***\end{tabular}} & \multicolumn{1}{c|}{\textbf{\begin{tabular}[c]{@{}c@{}}98.82\%\\ **\end{tabular}}} & \textbf{\begin{tabular}[c]{@{}c@{}}30.43\%\\ ****\end{tabular}} & \textbf{\begin{tabular}[c]{@{}c@{}}16.85\%\\ ***\end{tabular}} & \textbf{\begin{tabular}[c]{@{}c@{}}100.00\%\\ **\end{tabular}} \\ \hline
\multicolumn{1}{c|}{\begin{tabular}[c]{@{}c@{}}Compared with\\ Previous Solution\end{tabular}} & \textbf{} & \textbf{} & \multicolumn{1}{c|}{\textbf{}} & \textbf{\begin{tabular}[c]{@{}c@{}}24.79\%\\ ****\end{tabular}} & \textbf{\begin{tabular}[c]{@{}c@{}}6.32\%\\ **\end{tabular}} & \multicolumn{1}{c|}{\textbf{-}} & \textbf{\begin{tabular}[c]{@{}c@{}}22.10\%\\ ****\end{tabular}} & \textbf{\begin{tabular}[c]{@{}c@{}}-5.31\%\\ n.s.\end{tabular}} & \multicolumn{1}{c|}{\textbf{-}} & \textbf{\begin{tabular}[c]{@{}c@{}}26.17\%\\ ****\end{tabular}} & \textbf{\begin{tabular}[c]{@{}c@{}}6.53\%\\ **\end{tabular}} & \multicolumn{1}{c|}{\textbf{-}} & \textbf{\begin{tabular}[c]{@{}c@{}}27.74\%\\ ****\end{tabular}} & \textbf{\begin{tabular}[c]{@{}c@{}}7.94\%\\ **\end{tabular}} & \textbf{0.00\%} \\ \hline
\multicolumn{1}{c|}{\begin{tabular}[c]{@{}c@{}}Compared with\\ Ours w/o BC\end{tabular}} & \textbf{} & \textbf{} & \multicolumn{1}{c|}{\textbf{}} & \textbf{} & \textbf{} & \multicolumn{1}{c|}{\textbf{}} & \textbf{\begin{tabular}[c]{@{}c@{}}-3.58\%\\ n.s.\end{tabular}} & \textbf{\begin{tabular}[c]{@{}c@{}}-12.41\%\\ n.s.\end{tabular}} & \multicolumn{1}{c|}{\textbf{\begin{tabular}[c]{@{}c@{}}-99.94\%\\ n.s.\end{tabular}}} & \textbf{\begin{tabular}[c]{@{}c@{}}1.82\%\\ n.s.\end{tabular}} & \textbf{\begin{tabular}[c]{@{}c@{}}0.22\%\\ n.s.\end{tabular}} & \multicolumn{1}{c|}{\textbf{\begin{tabular}[c]{@{}c@{}}66.65\%\\ n.s.\end{tabular}}} & \textbf{\begin{tabular}[c]{@{}c@{}}3.91\%\\ *\end{tabular}} & \textbf{\begin{tabular}[c]{@{}c@{}}1.73\%\\ *\end{tabular}} & \textbf{\begin{tabular}[c]{@{}c@{}}100.00\%\\ n.s.\end{tabular}} \\ \hline
\multicolumn{1}{c|}{\begin{tabular}[c]{@{}c@{}}Compared with\\ Ours w/o RL\end{tabular}} & \textbf{} & \textbf{} & \multicolumn{1}{c|}{\textbf{}} & \textbf{} & \textbf{} & \multicolumn{1}{c|}{\textbf{}} & \textbf{} & \textbf{} & \multicolumn{1}{c|}{\textbf{}} & \textbf{\begin{tabular}[c]{@{}c@{}}5.21\%\\ n.s.\end{tabular}} & \textbf{\begin{tabular}[c]{@{}c@{}}11.24\%\\ **\end{tabular}} & \multicolumn{1}{c|}{\textbf{\begin{tabular}[c]{@{}c@{}}83.32\%\\ n.s.\end{tabular}}} & \textbf{\begin{tabular}[c]{@{}c@{}}7.22\%\\ *\end{tabular}} & \textbf{\begin{tabular}[c]{@{}c@{}}12.58\%\\ **\end{tabular}} & \textbf{\begin{tabular}[c]{@{}c@{}}100.00\%\\ n.s.\end{tabular}} \\ \hline
\multicolumn{1}{c|}{\begin{tabular}[c]{@{}c@{}}Compared with\\ Ours w/o DAgger\end{tabular}} & \textbf{} & \textbf{} & \multicolumn{1}{c|}{\textbf{}} & \textbf{} & \textbf{} & \multicolumn{1}{c|}{\textbf{}} & \textbf{} & \textbf{} & \multicolumn{1}{c|}{\textbf{}} & \textbf{} & \textbf{} & \multicolumn{1}{c|}{\textbf{}} & \textbf{\begin{tabular}[c]{@{}c@{}}2.13\%\\ **\end{tabular}} & \textbf{\begin{tabular}[c]{@{}c@{}}1.51\%\\ n.s.\end{tabular}} & \textbf{\begin{tabular}[c]{@{}c@{}}100.00\%\\ n.s.\end{tabular}}
\end{tabular}%
}
\end{table*}

\section{Experiment Results}
\label{sec:experiment_result}

In this section we design experiments to test the following research questions in both deterministic MPC and sampling-based MPC problem:
\begin{itemize}
    \item \textbf{RQ1} Does our warm-start policy reduce solver runtime in deterministic MPC problem? 
    \item \textbf{RQ2} Does our warm-start policy improve tracking performance in deterministic MPC problem?
    \item \textbf{RQ3} Does our warm-start policy generalize to unseen tracks in deterministic MPC problem?
    \item \textbf{RQ4} Does our warm-start policy generalize from deterministic MPC to sampling-based MPC problem?
\end{itemize}

\vspace{-0.3em}
\subsection{\RALedit{Deterministic} MPC}

We perform testing on both the training tracks (Fig.~\ref{fig:training_tracks}) and the challenging zero-shot tracks (Fig.~\ref{fig:zero_shot_tracks_complex}). The results are shown in Table~\ref{table:f1_results}. On both the training and zero-shot tracks, employing a warm-start policy trained through either offline learning or a combination of offline and online fine-tuning significantly reduces MPC optimization time and improves tracking accuracy. Besides, the warm-start policy trained with both offline and online fine-tuning achieves better MPC optimization time and tracking accuracy compared to the policy trained solely via offline BC, demonstrating the capability of our online fine-tuning algorithm in addressing the suboptimality and covariance shift problem in BC.

On the training tracks, compared to Ours w/o RL, the proposed method improves optimization time by $19.95\%$ and tracking accuracy by $30.60\%$. When compared to Ours w/o RL, Ours w/o DAgger achieves a $19.12\%$ improvement in tracking accuracy; however, the optimization time degrades by $22.08\%$, highlighting the necessity of incorporating the DAgger training loss during RL training.

On the zero-shot tracks, compared to Ours w/o RL, the proposed method improves optimization time by $21.63\%$ and tracking accuracy by $34.12\%$. Similarly, Ours w/o DAgger achieves a $30.35\%$ improvement in tracking accuracy, but optimization time degrades by $18.00\%$. This degradation pattern aligns with that on the training tracks, further highlighting the importance of integrating the DAgger training loss during RL training.

\begin{figure}[!h]
  \centering
  \includegraphics[width= 0.38\textwidth]{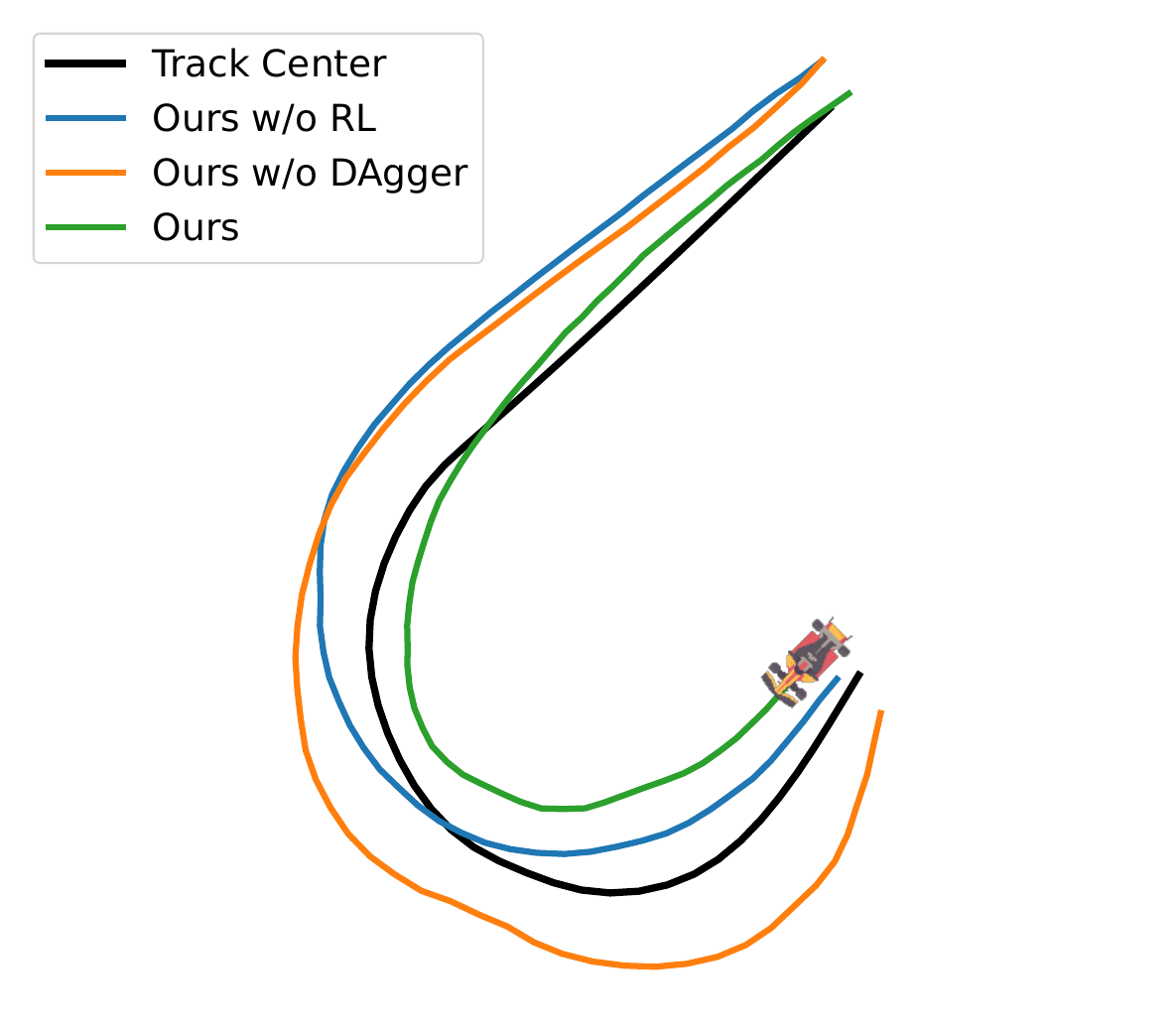}
  
  \caption{Visualization of optimized trajectories for our method and the baselines on a segment of the Nürburgring map.}
  \label{fig:trajectory_visualization}
\end{figure}

The results also reveal that initializing an MPC with either All Zeros or Previous Solution fails to complete laps on both the training and zero-shot tracks. The vehicle consistently deviates from the lane when facing any sharp turns. This limitation arises because the MPC often requires more iterations to optimize the control solution as the vehicle approaches the curves of the track. Since the real-time MPC only optimizes the solution for a maximum of 100 iterations at each step, the returned solution lacks the optimization necessary to guide the vehicle through sharp turns effectively. This underscores the necessity of a well-informed initial guess to minimize the number of optimization iterations required.

\RAL{We also relaxed the maximum MPC iterations to 200 and repeated the test. The results in Table~\ref{table:f1_results_200iter} show that our warm-start policy again achieves the greatest improvement in optimization time and the largest performance gains over the baselines, further supporting RQ1, RQ2, and RQ3.}

\RAL{A visualization of the optimized trajectories for our method and the baselines on a segment of the Nürburgring map is presented in Fig.~\ref{fig:trajectory_visualization} (MPC optimization iterations=100). The figure illustrates that using our method as a warm-start strategy enables the MPC to track the reference path more accurately by tracking the sharp more accurately.}



\vspace{-0.1em}
\subsection{Sampling-Based MPC}

We evaluate performance on the training maps (Fig.~\ref{fig:mppi_training_map}) and the zero-shot maps (Fig.~\ref{fig:mppi_testing_map}), with results summarized in Table~\ref{table:mppi_results}. Consistent with the findings for deterministic MPC, our proposed framework delivers the most significant performance improvements for the MPPI controller compared to all baselines. Notably, the improvements are more pronounced in the zero-shot environments, demonstrating the strong generalizability of the proposed method.


\subsection{Findings}
\vspace{-0.2em}

In summary, our empirical results support that:
\begin{itemize}
    \item \RALedit{Our proposed framework, trained offline or with online fine-tuning, outperform other non-learning-based baselines.}
    \item \RALedit{The two-phase training algorithm generalizes better on zero-shot tracks, achieving superior zero-shot performance.}
    \item \RALedit{Offline training effectively accelerates online fine-tuning, reducing training time to achieve strong performance.}
\end{itemize}
\vspace{-0.3em}

\section{LIMITATIONS AND FUTURE WORKS} 
\vspace{-0.2em}

While our proposed two-phase learning framework shows promising results in expediting optimization processes and enhancing control performance for robot control tasks, it also has several limitations for future research that merit consideration. \RAL{First, our experiments assume access to perfect state information.
Second, the current evaluation is limited to vehicle control. Future work could expand into higher-DOF robotic platforms such as manipulators or aerial robots.}

\vspace{-0.4em}
\section{CONCLUSIONS}
\vspace{-0.2em}

In this paper, we introduce a novel approach to accelerate MPC optimization by learning a warm-start policy. Our two-phase framework combines offline BC and online fine-tuning to provide better initial guesses for the MPC solver. Experiments on training and zero-shot tracks demonstrate the effectiveness of our approach in reducing optimization time without degrading MPC's performance. This integration of learning with MPC enhances the efficiency and applicability of trajectory optimization in dynamic systems. 
\vspace{-0.4em}

\bibliography{reference}
\bibliographystyle{IEEEtran}


\end{document}